\documentclass[journal]{IEEEtran}
\newcommand{\figref}[1]{{Fig.~\ref{#1}}}
\newcommand{\tabref}[1]{{Table.~\ref{#1}}}

\newcommand{\secref}[1]{Section.~\ref{#1}}
\usepackage{caption}
\newcommand{\figcaption}[1]{\def\@captype{figure}\caption{#1}}
\newcommand{\tblcaption}[1]{\def\@captype{table}\caption{#1}}
  \usepackage[pdftex]{graphicx}
\usepackage{here}

\usepackage{bm}
\usepackage{amsmath}
\usepackage{relsize}
\usepackage[subrefformat=parens]{subcaption}
\usepackage{siunitx}

\usepackage{placeins}

\usepackage{ifpdf}
\usepackage{cite}
\usepackage{algorithmic}

\usepackage{array}

\usepackage{url}
\begin{document}

\title{Design, Control, and Motion Strategy of TRADY: Tilted-Rotor-Equipped Aerial Robot With Autonomous In-flight Assembly and Disassembly Ability}

\author{Junichiro\;Sugihara,
        Takuzumi\;Nishio,
        Keisuke\;Nagato,
        Masayuki\;Nakao,
        and\;Moju\;Zhao% <-this % stops a space
\thanks{Junichiro\;Sugihara,\;Keisuke\;Nagato,\;Masayuki\;Nakao,\;and
        Moju\;Zhao\;(corresponding\;author)\;are\;with\;the\;Department\;of\;Mechanical-Engineering,\;The\;University\;of\;Tokyo,\;Bunkyo-ku,\;Tokyo\;113-8656,\;Japan\;(e-mail:j-sugihara@jsk.imi.i.u-tokyo.ac.jp;\;chou@jsk.imi.i.u-tokyo.ac.jp;\;nagato@hnl.t.u-tokyo.ac.jp;\;nakao@hnl.t.u-tokyo.ac.jp).}% <-this %
        % stops a space
\thanks{Takuzumi\;Nishio\;is\;with\;the\;Department\;of\;Mechano-Infomatics,\;The University\;of\;Tokyo,\;Bunkyo-ku,\;Tokyo\;113-8656,\;Japan\;(e-mail:nishio@jsk.imi.i.u-tokyo.ac.jp).}% <-this % stops a space
% \thanks{Manuscript received April 19, 2005; revised August 26, 2015.}
}
\maketitle

\begin{abstract}
Various types of aerial robots have been demonstrated in prior works
 with the intention of enhancing their maneuverability or manipulation
 capabilities. However, the problem remains in earlier researches that
 it is difficult to achieve both mobility and manipulation capability. This
 issue arises due to the fact that aerial robots with high mobility
 possess insufficient rotors to execute manipulation tasks, while
 aerial robots with manipulation ability are too large to achieve
 high mobility. To tackle this problem, we introduce in this article a
 novel aerial robot unit named TRADY. TRADY is a tilted-rotor-equipped
 aerial robot with autonomous in-flight assembly and disassembly
 capability. It can be autonomously combined and separated from another
 TRADY unit in the air, which alters the degree of control freedom of
 the aircraft by switching the control model between the under-actuated
 and fully-actuated models. To implement this system, we begin by
 introducing a novel design of the docking mechanism and an optimized
 rotor configuration. Additionally, we present the configuration of the
 control system, which enables the switching of controllers between
 under-actuated and fully-actuated modes in the air. We also include the
 state transition method, which compensates for discrete changes during
 the system switchover process. Furthermore, we introduce a new motion
 strategy for assembly/disassembly motion that incorporates recovery
 behavior from hazardous conditions. Finally, we evaluate the
 performance of our proposed platform through experiments, which
 demonstrated that TRADY is
 capable of successfully executing aerial assembly/disassembly motions
 with a rate of approximately 90\%. Furthermore, we confirmed that in the assembly state,
 TRADY can utilize full-pose tracking, and it can generate more than
 nine times the torque of a single unit. To the best of our knowledge,
 this work represents the first time that a robot system has been
 developed that can perform both assembly and disassembly while
 seamlessly transitioning between fully-actuated and under-actuated
 models.

\end{abstract}


% \fi
\IEEEpeerreviewmaketitle

\section{Introduction} \label{sec:introduction}
\IEEEPARstart{I}{n recent} years, aerial robots have undergone
significant development \cite{Floreano2015}, \cite{kumar2012} and have proven to be useful for various
practical applications. Compact aerial robots such as quadrotors with
terrain-independent mobility have enabled
\begin{figure}[H]
 \begin{center}
   \includegraphics[width=\columnwidth]{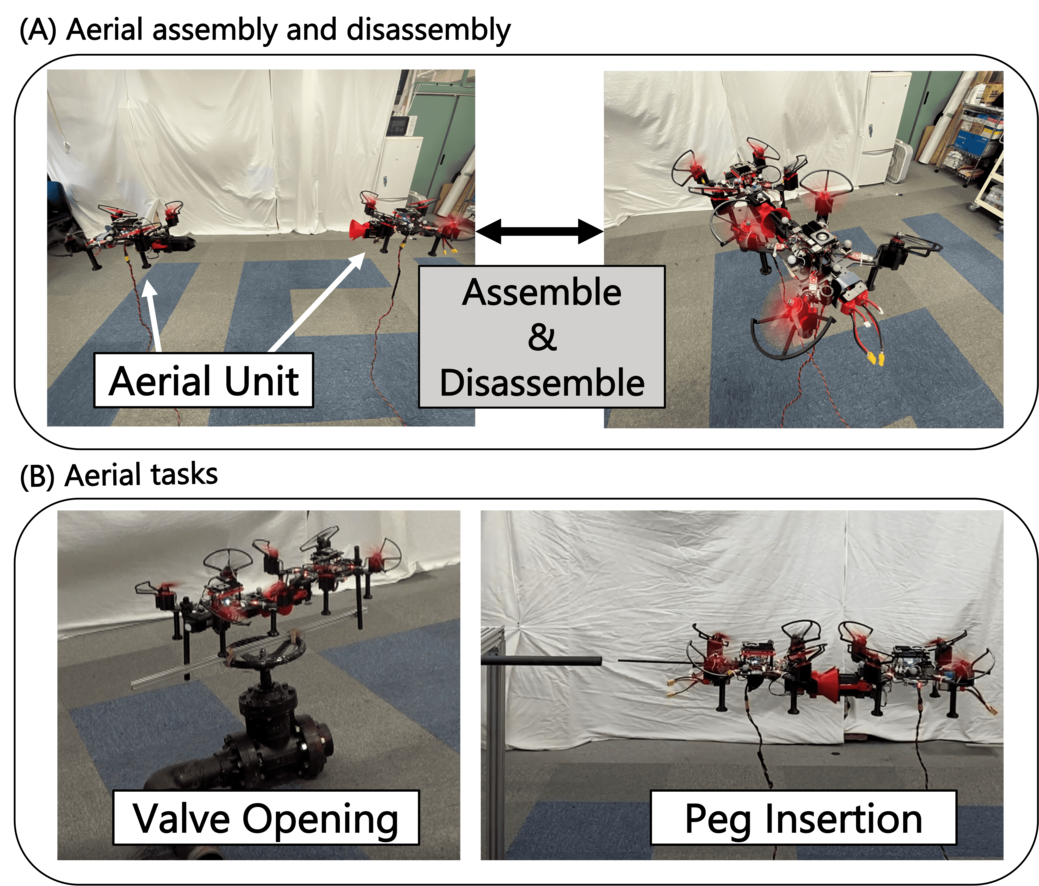}
 \end{center}
 \caption{The proposed robot platform TRADY: \textbf{T}ilted-\textbf{R}otor-Equipped Aerial Robot
 With Autonomous In-flight \textbf{A}ssembly and \textbf{D}isassembly
 Abilit\textbf{Y}. (A)The unitary state and
 the assembly state of TRADY. (B)Task execution capability of TRADY.}
   \label{figure:trady_main}
\end{figure}\noindent\\ \\ various autonomous applications such as cinematography \cite{Bonatti2020},
inspection \cite{Sewer}, disaster response \cite{Michael2012}, and
surveillance \cite{Doitsidis2012}. Furthermore, to enhance mobility, numerous designs that
enable robots to navigate through narrow spaces have been proposed, such
as quadrotors with morphing capabilities that allow their airframe to
shrink \cite{figure_8,morphing_quad, origami,n_zhao,Bucki}. On the other hand, there has been a growing demand for aerial
robots to possess manipulation capabilities, which require larger
controllable degrees of freedom (DoF) and available force and torque (wrench). As a
solution, more intricate and sizable aerial robots have been developed,
including multirotors equipped with over six tiltable rotors enabling
omni-directional control \cite{Voliro}, and arm-manipulator-installed multirotors
capable of interacting with the environment \cite{khamseh2018,Mellinger2011,Heredia2014}.\par
However, a significant issue in the conventional works is the inherent trade-off between
mobility and manipulation capability. To clarify, High-mobility
under-actuated aerial robots often lack the necessary controllable DoF
and available wrench to achieve manipulation tasks, whereas
fully-actuated robots with manipulation capability often lack
compactness. To address this problem, some researchers have
proposed a transformable multilink design \cite{hydrus,dragon,Lasdra} that enables robots to
navigate narrow spaces and perform manipulation tasks using their entire
body \cite{Zhao2022}. Nonetheless, these transformable aerial robots
require a significant number of additional
actuators, making them heavy, and even with transforming capability, it remains challenging
to achieve the same maneuverability as a small quadrotor in a narrow
space with three-dimensional complexity.\par
In this article, we present a solution to the aforementioned trade-off problem by introducing TRADY, a novel aerial robot platform that can perform aerial assembly
and disassembly motion, as depicted in \figref{figure:trady_main} (A), while changing
its controllable DoF and available wrench. In its unitary state, TRADY
is a compact under-actuated quadrotor, however, in its assembly state,
it becomes a fully-actuated octorotor, enabling it to execute aerial
manipulation tasks as depicted in \figref{figure:trady_main} (B).

\subsection{Related works}
\begin{figure}[!t]
 \begin{center}
   \includegraphics[width=0.8\columnwidth]{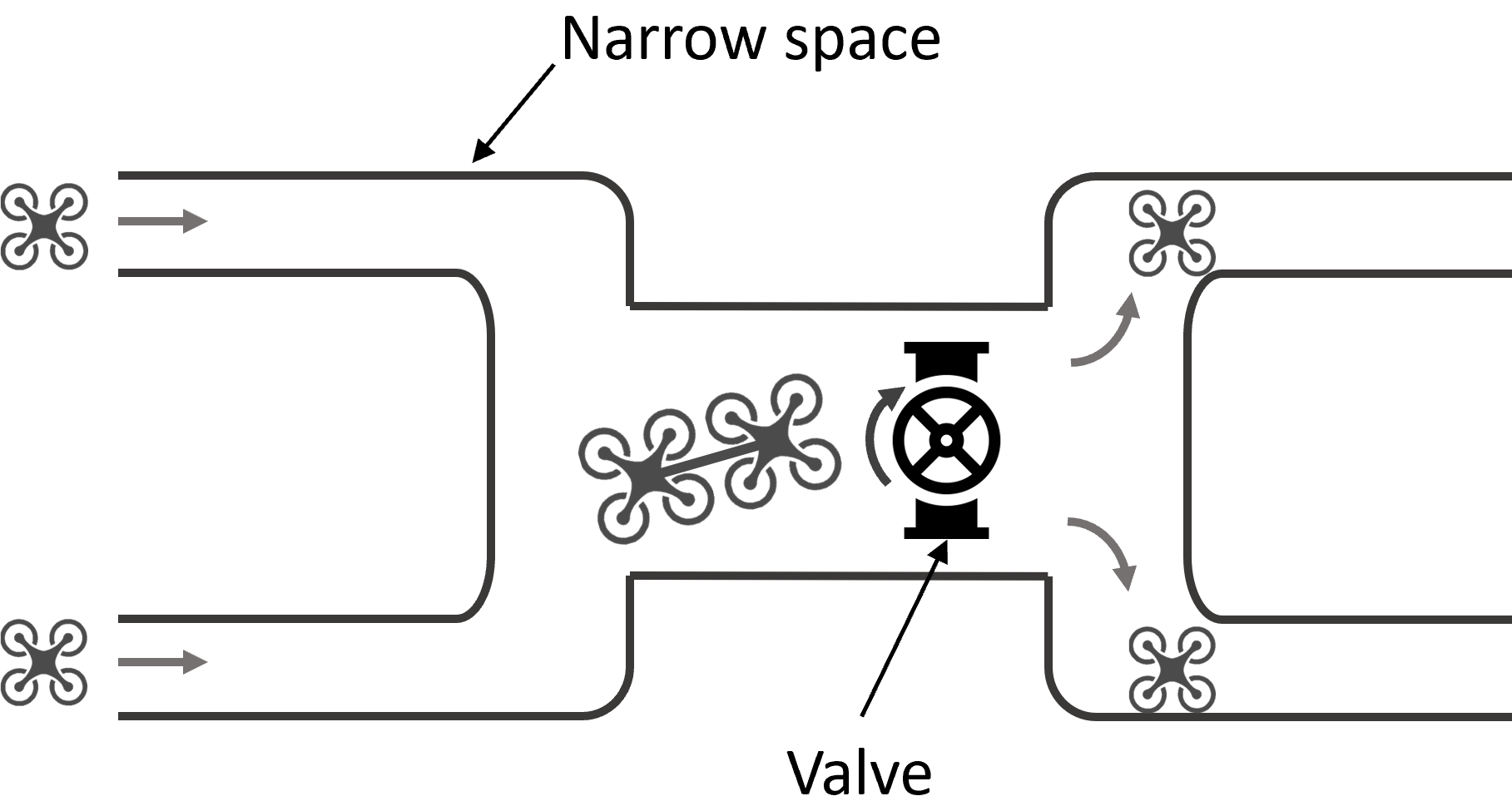}
 \end{center}
 \caption{Application example that can be realized with TRADY developed in this work: Valve opening and closing tasks in narrow and complex spaces.}
 \label{figure:trady_concept}
\end{figure}

TRADY is a type of modular aerial robot, and previous studies have
 developed several other modular aerial robots. In \cite{Naldi2015}, Naldi
 et al. introduce a single ducted-fan module, and in \cite{Granger}, a
 tetrahedron-shaped quadrotor module consisting of four fractal
 single-rotor submodules is presented. Furthermore, in \cite{Xu2021}, a
 modular quadrotor equipped with tilted rotors is proposed. These
 modular aerial robots can combine with one another and expand their
 degrees of freedom and wrench, however, \cite{Naldi2015,Granger,Xu2021} are currently not focused
 on self-assembly/disassembly. On the other hand, several aerial robots
 capable of self-assembly or self-disassembly have previously been developed. For
 instance, in \cite{Flightarray1}, \cite{Flightarray2}, a robot unit composed of wheels
 and rotors capable of docking on the ground and taking off afterwards
 is presented. Additionally, in \cite{Saldana2018}, a modular quadrotor with
 the capability of aerial self-assembly is proposed, followed by one
 with the ability of aerial self-disassembly proposed in
 \cite{moddessemble}. However, the issue that remains is that the robot
 presented in \cite{Flightarray1}, \cite{Flightarray2} is restricted to
 assembly/disassembly on the ground, and mid-air assembly/disassembly are unfeasible,
 whereas \cite{Saldana2018}, \cite{moddessemble} virtually specialize in
 either aerial assembly or aerial disassembly but not both. Futhermore, assembly or disassembly motion
 demonstrated in
 \cite{Flightarray1,Flightarray2,Saldana2018,moddessemble} are
 unaccompanied by the alternation of controllable DoF, implying that the
 robots are under the control of an under-actuated model in both the
 assembly and unitary states. The utilization instance
 of TRADY envisioned in this research comprises tasks illustrated in
 \figref{figure:trady_concept}. To accomplish this, movement in a narrow
 space must be carried out in the unitary state, and during manipulation
 tasks, the
 units must assemble to extend the controllable DoF and wrench, then
 return to the unitary state afterwards. Therefore, the compatibility of
 aerial self-assembly and self-disassembly, and the consequent alteration of the
 controllable DoF, remain crucial unachieved issues that are
 necessary for realizing TRADY. In this study, we tackle this issue
 from the perspectives of design, control, and motion strategy.\par

\par
Concerning the design of the self-docking mechanism, various designs
have been previously presented. For example, in \cite{Hara2014},
autonomous boats equipped with hook-string docking mechanisms are
introduced. In this system, the hooks of the male mechanism attach to
the loop of string on the female mechanism, and the female side winches
the loop of thread and the male mechanism together, thereby merging the
modules. While this mechanism works effectively on the water, the
coupling force is not sufficiently strong to be applicable to aerial
systems. On the other hand, in \cite{Saldana2018}, to achieve aerial docking, a
quadrotor is enclosed in a rectangular frame equipped with permanent
magnets. While this mechanism can construct a highly rigid structure, it
has the disadvantage of being unable to release the coupling once it has
been established. On the contrary, a quadrotor unit that can undock
through the torque generated by the unit itself is developed in
\cite{moddessemble}. However, to achieve undocking capability, the docking mechanism is
significantly downsized in comparison to the one used in \cite{Saldana2018}, and the
design is not suitable for aerial docking. Furthermore, as the docking
can be released through the torque output of a unit, the strength of the
mechanism becomes a bottleneck for available torque, making it arduous to execute
tasks that necessitate high torque. As an alternative method for
releasing the magnetic coupling, the utilization of elastic energy is
proposed in \cite{Yanagimura}. The mechanism proposed in \cite{Yanagimura} shortens the internal
spring of the mechanism using an actuator, and releases it to generate
restoring force, thereby releasing the magnetic coupling. In the case of
this mechanism, it is possible to release high rigidity coupling by
using a spring with a high spring constant and a powerful
actuator. However, pulling and detaching magnets in the translational
direction is an inefficient, resulting in the mechanism becoming
extremely large. Furthermore, with the docking mechanism that uses
magnets, in addition to the method of detachment, attention should be
paid to the magnetic interference from the external environment. When
the structure exposes the magnet to the outside as in [1], [2], although
normal flight is not a problem, magnetic interference from the
surroundings becomes a significant obstacle in manipulation tasks and
interaction with the external environment. Therefore, in this work, we introduce a novel
high-rigidity docking mechanism, which consists of a powerful magnetic mechanism that can be switched
on and off by a low-torque motor and a movable peg, thus enabling both
aerial docking and undocking to be achieved. Furthermore, we ensure that
the magnets are not exposed to the outside during flight through
appropriate structural design.\par

Regarding rotor configuration, in order to expand the controllable DoF
and available wrench through assembly and change the control model from
an under-actuated model to a fully-actuated model, it is necessary to
use tilted rotors. As mentioned earlier, \cite{Xu2021} has made it possible to
achieve fully-actuated model control by combining four quadrotor units
of different types with tilted rotors. In this case, the concept of the
actuation ellipsoid is used to determine the appropriate rotor
configuration. In this work, we propose a new rotor configuration
optimization method that focuses not only on maximizing the performance
in the assembly state, but also on achieving high-precision aerial
self-assembly/disassembly motion. Additionally, in order to achieve
simplicity in the system, TRADY acheive fully-actuated model by docking
two unit quadrotors with the same rotor configuration.\par

Subsequently, regarding the controller, TRADY utilizes distinct controllers for
the assembly and unitary states. The aircraft in the assembly state is a
typical fully-actuated multirotor, which can employ a conventional hexarotor control
method based on thrust allocation matrices, as presented in
\cite{Ryll2016}, \cite{Park2016}. By means of this method, TRADY in
the assembly state is capable of controlling six DoF, which consist of
forces and torques in all axes. On the other hand, TRADY in its unitary state
is a tilted quadrotor that is controlled as a under-actuated model. In this
case, the controllable DoF are four: the z-direction
force and torque on all axes, but it is not possible to control the x
and y-directional forces generated by the tilted rotors. Therefore, it is
necessary to suppress the generation of these horizontal forces through some
means. While in \cite{Xu2021},
which also uses a tilted quadrotor as a unit machine, the horizontal
forces are suppressed solely through design method, in this work, we focuses on suppressing the
horizontal forces through both design and control method. Therefore, LQI
control is adopted for unit control, explicitly minimizing the
horizontal forces generated during flight. Another important point to
consider in the control of TRADY is the switching between under-actuated
and fully-actuated models during aerial assembly/disassembly motion. When the control
model is discretely changed during flight, the control of the vehicle
becomes unstable (e.g., the vehicle suddenly rising or falling) due to
the influence of model errors. The elimination of this instability is
crucial for achieving stable assembly/disassembly. Therefore, in this
study, we introduce a novel method to eliminate this instability by applying
our own transition processing during the control model switching. \par
Finally, regarding motion strategy, several previous studies have
presented strategies for self-assembly of robots. For instance, in
\cite{Hara2014}, a path planning method based on global optimization is employed to
enable self-assembly of boats on water surfaces, while \cite{Saldana2018} utilizes vehicle
guidance via gradient method for quadrotor self-assembly in
mid-air. These strategy are efficacious in circumstances
where the high positional accuracy of unit vehicles can be
maintained during the assembly motion. Nonetheless, when it comes to the
aerial assembly motion of unit vehicles that are furnished with tilted rotors,
as employed in this study, the thrust of each unit interferes with one
another, resulting in unstable position control as the two units
approach. Consequently, in this study, we assume the existence of
positional errors and suggests a motion strategy that iteratively
approaches until the assembly is successfully achieved,  while avoiding
dangerous situations. This facilitates an autonomous and reliable
assembly motion. Additionally, we extend the strategy to encompass
self-disassembly, providing a comprehensive approach.\par
% \IEEEpubidadjcol
To sum up, the main contribution of this work can be summarized as
follows;
\vspace{-2pt}
 \begin{enumerate}
 \item We propose the design of docking mechanisms that combines strong
       coupling and easy separation.
 \item We present the optimized rotor configuration to achieve controllability
       with under-actuated model in the unitary state and with
       fully-actuated model in the assembly state. Moreover, this rotor
       configuration also takes into account the stability improvement
       of aerial assembly/disassembly motion.
 \item We develop a control system that allows for switching between
       control models and includes transition processing to compensate
       for control instability during model switching.
 \item We introduce the motion strategy that enables autonomous and
       stable aerial assembly/disassembly motion even in
       situations where position control is unstable.
 \end{enumerate}
 \vspace{-2pt}
Although we focus on the evaluation with two units, our methodology can be easily applied to more
units by installing both male and female docking mechanisms in a single
unit. Furthermore, to the best of our knowledge, this is the first time
to achieve the both self-assembly and self-dessembly with the same robot
platform in an autonomous manner, and also achieved the manipulation
task by switching between under-actuated and fully-actuated models.
\subsection{Notation}
From this section, nonbold lowercase symbols (e.g., $m$) represent
scalars, nonbold uppercase symbols (e.g., $R$) represent sets or linear spaces, and
bold symbols (e.g., $\bm{u}, \bm{Q}$) represent vectors or
matrices. Superscripts (e.g, ${}^{\lbrace{CoG}\rbrace}\bm{p}$)
represent the frame in which the vector or matrics is expressed, and
subscripts represent the target frame or an axis, e.g.,
${}^{\lbrace\ W\rbrace}\bm{r}_{\lbrace CoG \rbrace}$ represents a
vector point from ${\lbrace W\rbrace}$ to ${\lbrace CoG\rbrace}$
w.r.t. ${\lbrace W\rbrace}$, whereas $u_{x}$ denotes the $x$ component
of the vector $\bm{u}$.
\subsection{Organization}
The remainder of this paper is organized as follows. The mechanical
design including the design of docking mechanisms is presented in
\secref{sec:mech_design}, and the modeling of our
robot and optimized rotor configuration is introduced in
\secref{sec:rotor_config}. Next, the flight control and model switching method are
presented in \secref{sec:control}, followed by motion strategy in
\secref{sec:motion_strategy}. We then show the experimental result of
trajectory following flights, aerial assembly and disassembly, and
object manipulation tasks in \secref{sec:experiment} before concluding
in \secref{sec:conclusion}.

\section{Mechanical Design}\label{sec:mech_design}
In this section, we present the mechanical design of the proposed robot,
TRADY. Initially, we provide an overview of the entire robot design and
subsequently expound upon the design of the docking mechanisms.
\subsection{Entire Robot Design}\label{subsec:whole_design}
\begin{figure}[!t]
 \begin{center}
   \includegraphics[width=\columnwidth]{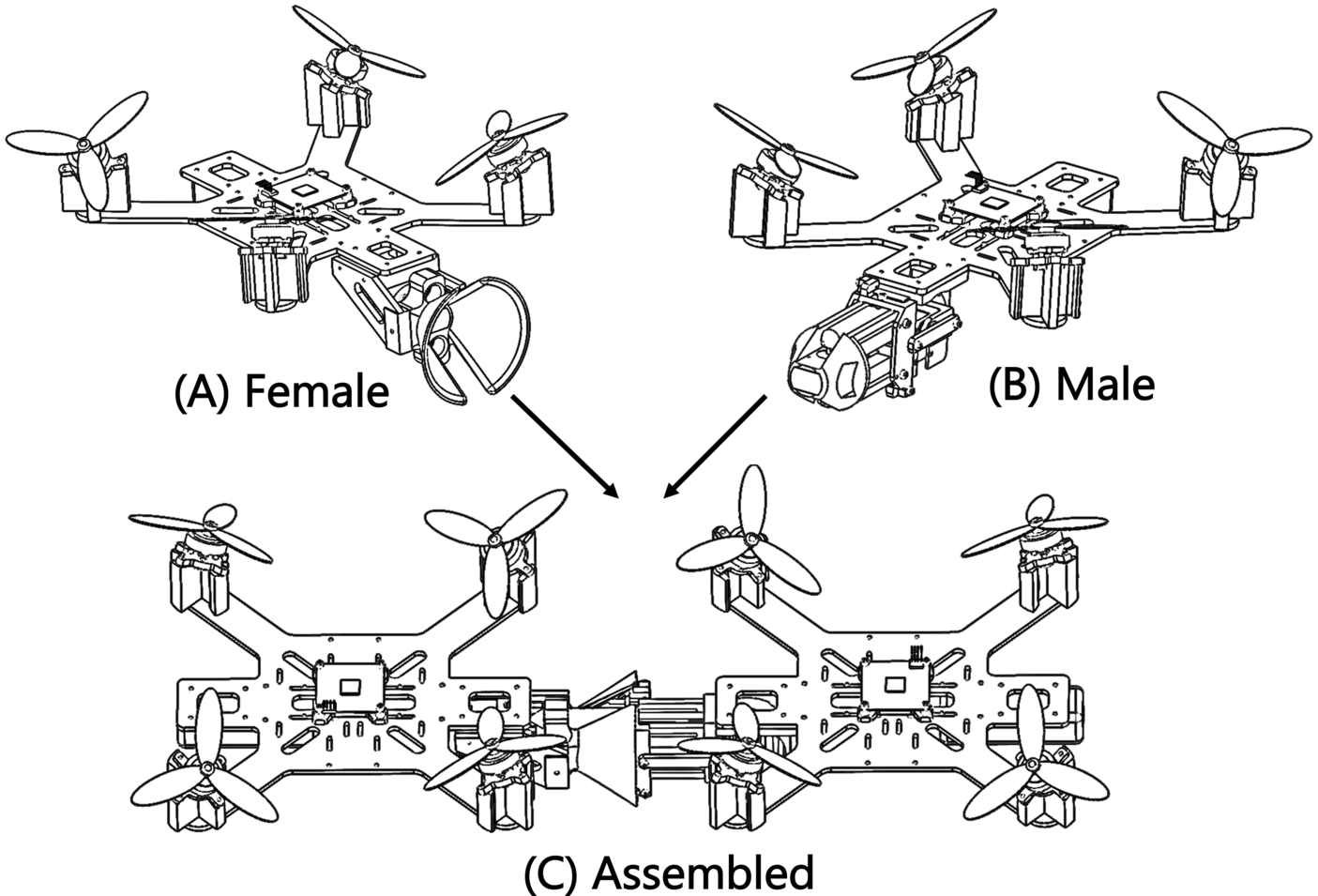}
 \end{center}
 \caption{TRADY's entire design. (A)Female unit. (B)Male
 unit. (C)Assembled state.}
   \label{figure:whole_design}
\end{figure}

In this study, the minimum unit comprising TRADY is designed as a
quadrotor unit. Each unit is equipped with a common rotor configuration,
as well as either a male-side or female-side the mechanism at the
same position. Therefore, the overall structure of the device is
depicted in \figref{figure:whole_design}.
\subsection{Docking Mechanism}\label{subsec:docking}
The prerequisites for the design of a docking mechanism are twofold:
firstly, it should be capable of accomplishing both docking and
undocking operations in the air, and secondly, it should possess
sufficient rigidity to execute high-load tasks. Therefore, we propose a
methodology of coupling with a permanent magnet and movable pegs.
\subsubsection{Female side}
\begin{figure}[!t]
 \begin{center}
   \includegraphics[width=0.8\columnwidth]{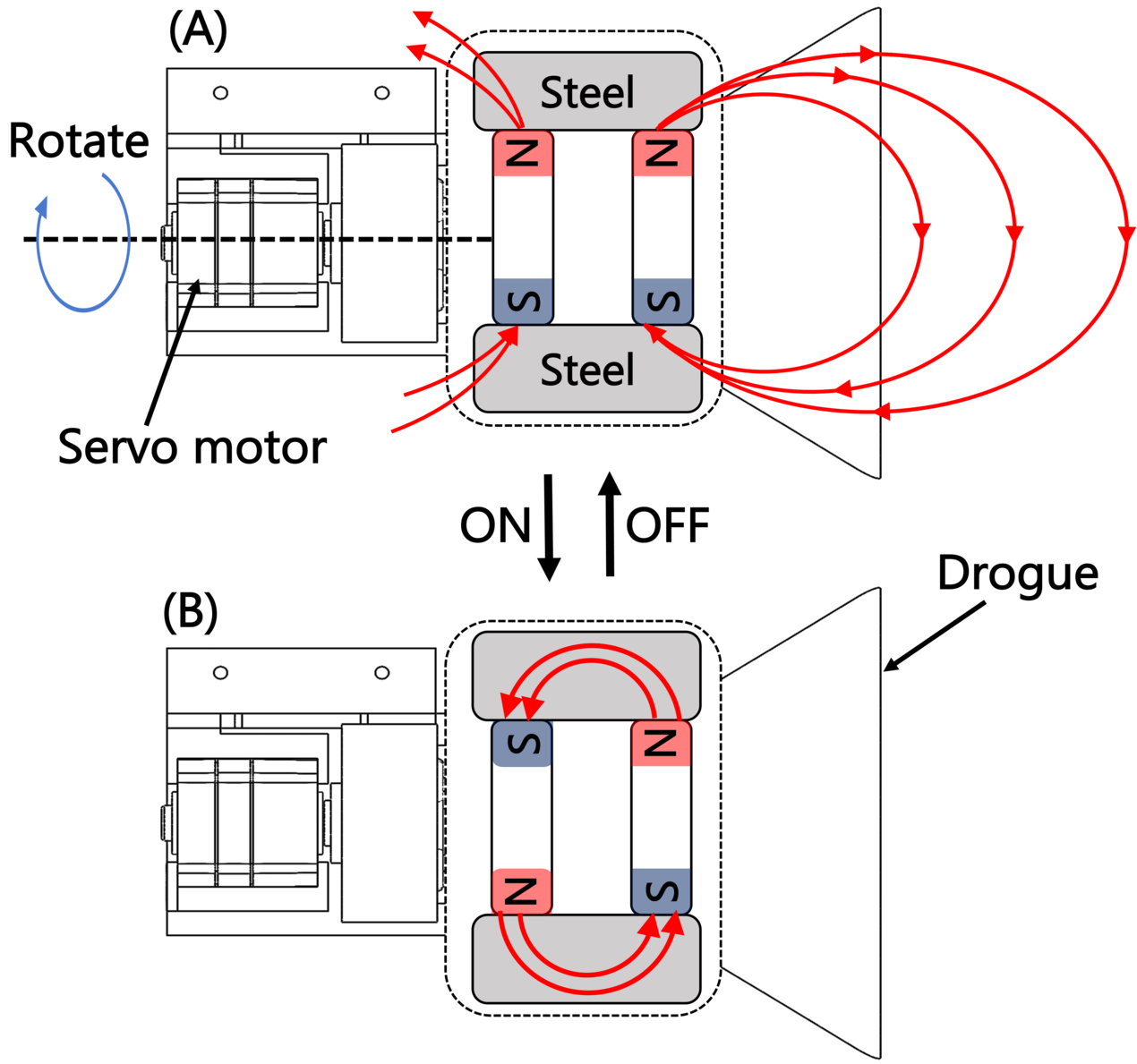}
 \end{center}
 \caption{How the female mechanism works. In (A), since the magnetic force
 lines of force extend beyond the system, magnetic forces acts on the
 outside. In (B), magnetic force does not act on the outsinde because
 the magnetic lines of force circulate within the system. Note that the
 size of the switchable magnet is exaggerated in the depiction.}
   \label{figure:female_mech}
\end{figure}

As shown in \figref{figure:female_mech}, the main components of the female mechanism are a
permanent magnet equipped with a magnetic switching mechanism and a
mechanism called a drogue, which compensates for positioning
errors. \par
The on-off switching of the magnetic force is realized by the principle
illustrated in the cut model in \figref{figure:female_mech}. In this mechanism, it is
possible to interrupt the magnetic force by rotating one of the two
permanent magnets \SI{180}{deg} using a servo motor. Since the two magnets do not directly touch each other, the force that
hinders the rotation of the magnet is only friction with the housing,
and rotation is easily possible with a low-torque servo motor. Through
the use of this magnet, restraint in the translational direction of the
docking mechanism is achieved.\par
Furthermore, the drogue attached to the tip of the female mechanism is a
mechanism adopted in autonomous aerial refueling systems for fighter
aircraft such as \cite{Tandale}, \cite{FRAVOLINI2004611},
which compensates for control errors of the size corresponding to the
diameter of the mechanism. In this study, the radius of the drogue was
set to \SI{5}{cm} in order to absorb errors within
$\pm$\SI{2.5}{cm}. Additionally, the gradient inside the drogue was
empirically determined to be approximately \SI{35}{deg}. The drogue not
only compensates for position errors, but also has two additional
functions. One is to improve rigidity by increasing the contact area,
and the other is to prevent the influence of external magnetic forces by
covering the magnets.
\subsubsection{Male side}
\begin{figure}[!t]
 \begin{center}
   \includegraphics[width=0.7\columnwidth]{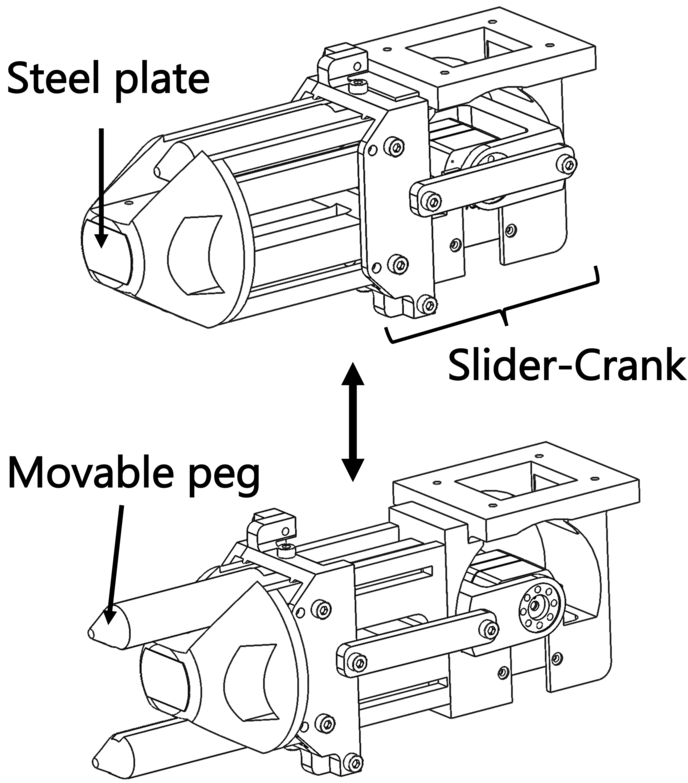}
 \end{center}
 \caption{How the male mechanism works. The pegs are driven by the
 slider-crank mechanism and inserted into female mechanism.}
   \label{figure:male_mech}
\end{figure}

As shown in \figref{figure:male_mech}, the male mechanism is composed of an steel plate
for magnetic attraction and movable pegs. The movement of the peg is
achieved using a slider-crank mechanism powered by a servo motor, and
during docking, the peg is inserted into the receptor of the female
mechanism to provide confinement in the bending direction. The tip of
the mechanism is designed to closely adhere to the drogue of the female
mechanism. \par
The overall size of the docking mechanism is determined
based on the clearance that should be maintained between the rotors in
the assembly state.

\section{Modeling and Rotor Configuration}\label{sec:rotor_config}
In this section, we introduce the kinematics and dynamics model of our
robot in the first place, followed by the thrust allocation. Then, we present the method of optimization of
rotor configuration.
\subsection{Modeling}

\begin{table}[!t]
  \renewcommand{\arraystretch}{1.3}
  \caption{Definition of physical quantities for multirotor.}
  \centering
 \begin{tabular}{|c|c|}
  \hline
  Symbol&Definition \\
  \hline
  $n$& Number of rotors\\
  $m$& Mass of body\\
  $\bm{I}$& Moment of inertia ob body\\
  ${}^{\lbrace\bm{W}\rbrace}$& World coordinates \\
  ${}^{\lbrace\bm{CoG}\rbrace}$& CoG coordinates \\
  $\bm{r}$& Position of CoG\\
  $\bm{R}$& Rotation matrics of body\\
  $\bm{\alpha} =\lbrack\begin{array}{ccc}
                  \theta&\phi &\psi \\
                       \end{array}\rbrack$ & Euler angles (roll, pitch, yaw)\\
  $\bm{p_{i}}$& Position of i-th rotor\\
  $\bm{u_{i}}$& Direction of i-th rotor( $\|\bm{u_{i}} \| = 1 $)\\
  $\bm{\lambda} = ^t\!\lbrack \lambda_{1},\lambda_{2},\cdots,\lambda_{N}\rbrack$& Thrust\\
  $\bm{\sigma} = \lbrack \sigma_{1},\sigma_{2},\cdots,\sigma_{N}
  \rbrack$& Counter torque of each rotor\\
  $\bm{f}$& Resultant force\\
  $\bm{\tau}$& Resultant torque\\
  $\bm{\omega}$&Angular velociy\\
  $\bm{g} = \lbrack\begin{array}{ccc}
                  0&0&g \\
                       \end{array}\rbrack$& Acceleration ofGravity\\
  \hline
 \end{tabular}
 \label{table:multirotorparam}
\end{table}
\renewcommand{\arraystretch}{1.0}

\begin{figure}[!t]
 \begin{center}
  \begin{minipage}[b]{0.6\columnwidth}
   \includegraphics[width=\columnwidth]{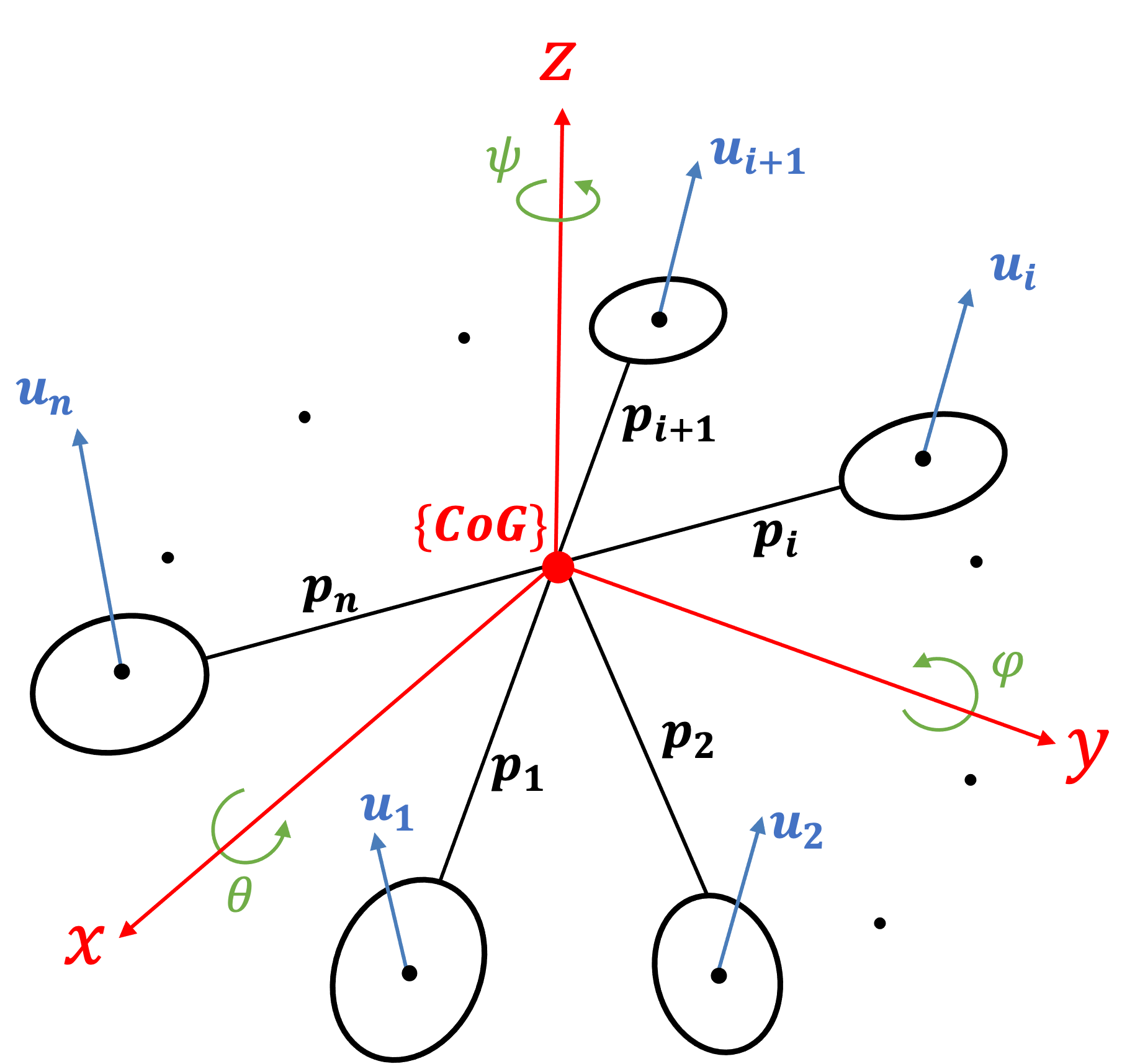}
   \vspace{+5pt}
   \centering{(A)}
  \end{minipage}
  \hspace{0.05\columnwidth}
   \begin{minipage}[b]{0.6\columnwidth}
    \includegraphics[width=\columnwidth]{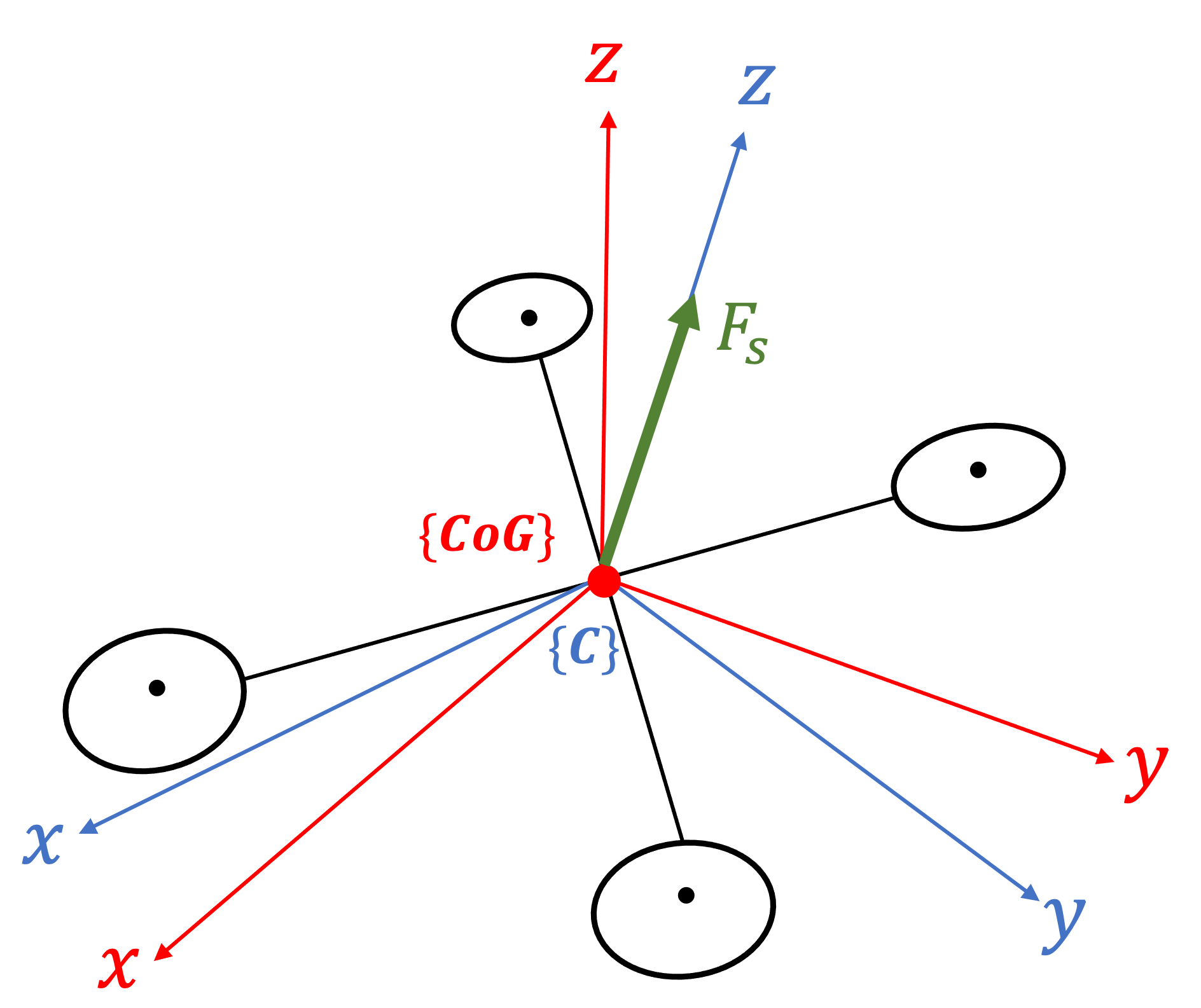}
    \vspace{+5pt}
  \centering{(B)}
  \end{minipage}
 \end{center}
 \caption{(A)Multi-rotor kinematics model with any number of rotors. The
 number of rotors is denoted by $n$; in this study, $n=8$ in the assemble
 state and $n=4$ for a unit. The frame ${\lbrace CoG \rbrace}$ have the
 origin at the center-of-mass of the whole model and z-axis is oriented
 vertically to airframe.
 (B)Relationship between ${\lbrace CoG \rbrace}$ and ${\lbrace C
 \rbrace}$. $\bm{F_{z}}$ in the figure is the resultant rotor thrusts
 force when the robot is hovering.}
 \label{figure:multirotor_model}
\end{figure}

The kinematics model of TRADY is depicted in
\figref{figure:multirotor_model}(A) and each quantities are difined as
shown in \tabref{table:multirotorparam}. Since this model can be applied
to multirotors with any number of rotors, the model of TRADY is
represented by \figref{figure:multirotor_model} with $n = 4$ in the unitary state, and with $n = 8$
in the assembly state.\par
Based on this kinematics model, the wrench-force
${}^{\lbrace{CoG}\rbrace}\bm{f}$ and torque
${}^{\lbrace{CoG}\rbrace}\bm{\tau}$ )  can be written as

\begin{equation}\label{equation:force}
 {}^{\lbrace{CoG}\rbrace}\bm{f} = \sum\limits_{i=1}^{n}\lambda_{i}{}^{\lbrace{CoG}\rbrace}\bm{u}_{i},
\end{equation}
\begin{equation}\label{equation:torque}
 {}^{\lbrace{CoG}\rbrace}\bm{\tau} = \sum\limits_{i=1}^{n}\lambda_{i}{}^{\lbrace{CoG}\rbrace}\bm{p}_{i}\times {}^{\lbrace{CoG}\rbrace}\bm{u}_{i}+\bm{\sigma}{}^{\lbrace{CoG}\rbrace}\bm{f},
\end{equation}
where ${\lbrace{CoG}\rbrace}$ is the frame that have the origin at
the center-of-mass of the body.
From \eqref{equation:force} and \eqref{equation:torque}, the
translational and rotational dynamics of a multirotor unit are given by the Newton-Euler
equation as followings:
\begin{equation}\label{equation:translation}
 M{}^{\lbrace W\rbrace}\ddot{\bm{r}}_{\lbrace CoG \rbrace} =
 {}^{\lbrace W\rbrace}\bm{R}_{\lbrace CoG \rbrace}{}^{\lbrace{CoG}\rbrace}\bm{f}+
 \left(\begin{array}{c}
  0 \\
        0 \\
        -Mg
       \end{array}
 \right),
\end{equation}
\begin{equation}\label{equation:rotation}
\begin{split}
 {}^{\lbrace{CoG}\rbrace}\bm{I}{}^{\lbrace{CoG}\rbrace}\dot{\bm{\omega}}
  &=
 {}^{\lbrace{CoG}\rbrace}\bm{\tau}+\bm{\sigma}{}^{\lbrace{CoG}\rbrace}\bm{f}\\
 &\quad -
 {}^{\lbrace{CoG}\rbrace}\bm{\omega}\times{}^{\lbrace{CoG}\rbrace}\bm{I}{}^{\lbrace{CoG}\rbrace}\bm{\omega},
\end{split}
\end{equation}
where ${\lbrace W\rbrace}$ frame represent the world coordinate system.
Then, using \eqref{equation:force} and
\eqref{equation:torque}, allocation from the thrust force $\bm{\lambda}$
to the resultant wrench can be given by following:
\begin{equation}\label{equation:allocation}
 \left(\begin{array}{c}
  {}^{\lbrace{CoG}\rbrace}\bm{f} \\
        {}^{\lbrace{CoG}\rbrace}\bm{\tau}
       \end{array}
 \right)
 =
 \left(\begin{array}{c}
  \bm{Q_{tran}}\\
  \bm{Q_{rot}}
       \end{array}
        \right)
 \bm{\lambda}
 =
 \bm{Q}\bm{\lambda},
\end{equation}
where
\begin{equation}\label{equation:forceallo}
\bm{Q_{tran}} = \left( {}^{\lbrace{CoG}\rbrace}\bm{u}_{1},{}^{\lbrace{CoG}\rbrace}\bm{u}_{2},\cdots,{}^{\lbrace{CoG}\rbrace}\bm{u}_{n} \right),
\end{equation}
\begin{equation}\label{equation:torqueallo }
\bm{Q_{rot}} = \left({}^{\lbrace{CoG}\rbrace}\bm{v}_{1},{}^{\lbrace{CoG}\rbrace}\bm{v}_{2},\cdots,{}^{\lbrace{CoG}\rbrace}\bm{v}_{n} \right),
\end{equation}
\begin{equation}\label{equation:defin_v}
{}^{\lbrace{CoG}\rbrace}\bm{v}_{i} = {}^{\lbrace{CoG}\rbrace}\bm{p}_{i}\times {}^{\lbrace{CoG}\rbrace}\bm{u}_{i}.
\end{equation}
Note that the second term in \eqref{equation:torque} is omitted for the
remainder of the analysis because it is one order of magnitude smaller
than the first term in general.
\subsection{Thrust Allocation}
\subsubsection{In the Assembly State}
In the assembly state, TRADY is fully-actuated, and the allocation matrics $\bm{Q}$
is full-rank. Therefore, we can gain MP pseudo-inverse
matrics $\bm{^{\#}Q}$.
Given a desired wrench, the target thrust can be computed by following:
\begin{equation}\label{equation:Qinv}
   \bm{\lambda}
    \:=\:
    \bm{^{\#}Q}
 \left(\begin{array}{c}
  {}^{\lbrace{CoG}\rbrace}\bm{f} \\
        {}^{\lbrace{CoG}\rbrace}\bm{\tau}
       \end{array}
 \right).
\end{equation}
\subsubsection{In the Unitary State}
In the unitary
state, $\bm{Q} \in R^{6\times 4}$ is rank deficient, and we need to adopt under-actuated model. As the conventional method, the control targets for
under-actuated model are $f_{z}$, $\tau_{x}$, $\tau_{y}$, and
$\tau_{z}$. Applying the control method illustrated in \secref{sec:control}, we can
achieve the hovering of a quadrotor unit with these four inputs.
In the case of a quadrotor with non-tilted rotors, because it does not produce
$f_{x}$ and $f_{y}$, desired thrusts are easily calcurated as following:
\begin{equation}\label{equation:Qinv_quad}
   \bm{\lambda}
    \:=\:
    \bm{^{\#}\left(^{quad}Q\right)}
 {\:}^{t}\lbrack\begin{array}{cccc}
  f_{z} &
  \tau_{x} &
  \tau_{y} &
  \tau_{z}
       \end{array}
 \rbrack,
\end{equation}
where
\begin{equation}\label{equation:Qinv_unit}
   \bm{^{quad}Q}
    \:=\:
    \left(
     \begin{array}{c}
  \bm{Q_{tran,z}}\\
  \bm{Q_{rot}}
     \end{array}
        \right) \in R^{4\times 4} ,
\end{equation}
and $\bm{Q_{tran, z}} \in R^{1 \times 4}$ is the third row vector
of $\bm{Q_{tran}}$.\par
However, TRADY unit is a tilted quadrotor that produces translation forces
$f_{x}$, $f_{y}$  and these are uncontrollable with
\eqref{equation:Qinv_quad}. Therefore,  assuming the existence of
the static thrust $\bm{\lambda_{s}}$ that enables hovering while suppressing the
generation of $f_{x}$ and $f_{y}$, we introduce a method to apply
\eqref{equation:Qinv_quad} to the thrust
allocation of tilted quadrotor by utilizing $\bm{\lambda_{s}}$. \par
In \cite{singularity}, Zhao et al. use a tilted
coodinate system in order to obtain $\bm{\lambda_{s}}$ and we apply this
idea to our work. Now, we introduce a new coodinate system ${\lbrace{C}\rbrace}$ that has
the origin at the center-of-mass of body as shown in \figref{figure:multirotor_model}(B). Furthermore, we define ${\lbrace{C}\rbrace}$ to fulfill the following conditions:
\begin{equation}\label{equation:force_condition}
 \bm{Q_{tran}}'\bm{\lambda_{s}} = -m\bm{g},
\end{equation}
\begin{equation}\label{equation:torque_condition}
 \bm{Q_{rot}}'\bm{\lambda_{s}} = \bm{0},
\end{equation}
where $\bm{Q_{tran}}'$, $\bm{Q_{rot}}'$ are allocation matrices defined
in ${\lbrace{C}\rbrace}$.\\
\eqref{equation:force_condition}, \eqref{equation:torque_condition}
indicate that the direction of the resultant force due to
$\bm{\lambda_{s}}$ coincides with the z-axis of
${\lbrace{C}\rbrace}$. In other word, by controlling the robot's attitude
so that ${\lbrace{C}\rbrace}$ is horizontal, it becomes possible to
hover the robot without generating excess forces in the horizontal
direction. In this case, by using $\bm{Q_{tran}}'$ and $\bm{Q_{rot}}'$,
thrust allocation can be achieved through \eqref{equation:Qinv_quad}, similar to that
of a non-tilt quadrotor. Therefore, we focus on deriving
$\bm{Q_{tran}}'$, $\bm{Q_{rot}}'$, and ${\lbrace{C}\rbrace}$.\par
Since both ${\lbrace{C}\rbrace}$ and ${\lbrace{CoG}\rbrace}$ have
origins at the center-of-mass, from \eqref{equation:torque_condition},
following is also satisfied:
\begin{equation}\label{equation:torquecondition2}
 \bm{Q_{rot}}\bm{\lambda_{s}} = 0.
\end{equation}
Additionally, regarding translation, following is valid from
\eqref{equation:force_condition}:
\begin{equation}\label{equation:forcecondition2}
 \| \bm{F_{z}} \| = mg,
\end{equation}
where $\bm{F_{z}} =  \bm{Q_{trans}}\bm{\lambda_{s}}$.
Then, we difine the rotation matrics $\bm{R_{C}}$ that satisfies:
\begin{equation}\label{equation:define_rc}
 \bm{R_{C}}\bm{F_{z}} ={}^{t}\lbrack
                                            \begin{array}{ccc}
                                             0&
                                             0&
                                             mg
                                            \end{array}
                                           \rbrack.
\end{equation}
Integrating \eqref{equation:torquecondition2},
\eqref{equation:forcecondition2}, and \eqref{equation:define_rc} we can gain follows:
\begin{equation}\label{equation:semifinal}
 \bm{R_{C}}
 \left(
  \begin{array}{c}
   \bm{Q_{tran}}\\
   \bm{Q_{rot}}
  \end{array}
        \right)
 \bm{\lambda_{s}}
 =
 {}^{t}\lbrack\begin{array}{cccccc}
  0&
        0&
        mg&
        0&
        0&
        0
       \end{array}\rbrack .
\end{equation}
Comparing
\eqref{equation:force_condition} and \eqref{equation:torque_condition} with
\eqref{equation:semifinal}, our discussion results in followings:
\begin{equation}\label{equation:final1}
\bm{R_{C}} \bm{Q_{tran}}=\bm{Q_{tran}}' ,
\end{equation}
\begin{equation}\label{equation:final2}
 \bm{R_{C}} \bm{Q_{rot}}=\bm{Q_{rot}}' .
\end{equation}
Because \eqref{equation:final1}, \eqref{equation:final2} mean that $\bm{R_{C}}$
is the rotation matrics that maps ${\lbrace{CoG}\rbrace}$ to
${\lbrace{C}\rbrace}$, the conversion from ${\lbrace{CoG}\rbrace}$ to
${\lbrace{C}\rbrace}$ is easily calcurated with
$\bm{R_{C}}$. Furthermore, \eqref{equation:force_condition} to \eqref{equation:final2} suggest that if there exists
a $\bm{Q}$ that satisfies \eqref{equation:force_condition} and
\eqref{equation:torque_condition} for a given $\bm{\lambda_{s}}$, it is
possible to hover tilted quadrotor using that $\bm{\lambda_{s}}$.
\subsection{Optimized Rotor Configuration}
The TRADY rotor configuration must meet three prerequisites. Firstly, it
must enable fully-actuated model control in the assembly
state. Secondly, it should allow for under-actuated model control in the
unitary state. Finally, it must attain the necessary flight properties
required for aerial assembly/disassembly motion. Therefore, we propose a methodology to
optimize the rotor configuration that satidsfies these conditions.\par
\subsubsection{Fully-actuated Model Controllability}
Initially, we propose a technique for achieving fully-actuated model
control when TRADY is in the assembly state. As described in Equation 1,
the full rank of matrix $\bm{Q}$ is equivalent to fully-actuated
model control, as $\bm{Q}$ can map L to an arbitrary wrench in real
space if it is full rank. However, each rotor has a limit of thrust
force, which can lead to instability of control due to weak resultant
force or torque in certain directions. Thus, rather than directly
employing algebraic methods to make $\bm{Q}$ full rank, we seek to
maximize the available force and torque region to ensure translational
and rotational controllability in all axes. To accomplish this, a
concept of feasible control force convex polyhedron
$\mathcal{V}_{\mathcal{F}}$, and torque convex polyhedron
$\mathcal{V}_{\mathcal{T}}$ were introduced by \cite{Convex}. These are
defined as follows:
\begin{equation}\label{equation:force_convex}
\mathcal{V}_{\mathcal{F}}\left(\bm{U}\right):=\left\{{}^{\lbrace{CoG}\rbrace}\bm{f} \in R^{3} \mid 0 \leq \lambda_{i} \leq \lambda_{\max }\right\},
\end{equation}
\begin{equation}\label{equation:torque_convex}
 \mathcal{V}_{\mathcal{T}}\left(\bm{U}\right):=
  \left\{
   {}^{\lbrace{CoG}\rbrace}\bm{\tau} \in {R}^{3} \mid 0\leq \lambda_{i}
   \leq \lambda_{\max} \right\} ,
\end{equation}
where the set of rotor direction vectors $\bm{U}$ is defined as
$\bm{U} = \lbrack\begin{array}{ccc}
 \bm{u_{1}}&\cdots &\bm{u_{8}} \\
                 \end{array}\rbrack$. In addition, ${}^{\lbrace{CoG}\rbrace}\bm{f}$,
${}^{\lbrace{CoG}\rbrace}\bm{\tau}$ are defined in
\eqref{equation:force}, \eqref{equation:torque} and maximum thrust for
each rotor in TRADY is denoted by $\lambda_{\max}$, while the minimum
thrust is established at 0, as we utilize unidirectional rotors.\par
Then,  we define the values for the guaranteed minimum control force,
denoted as $f_{min}$, and the corresponding torque, represented by
$\tau_{min}$, in accordance with the subsequent equations are being
satisfied:
\begin{equation}\label{equation:f_min}
 \| {}^{\lbrace{CoG}\rbrace}\bm{f} - m \bm{g} \| \le f_{min}\: \Rightarrow\:
  {}^{\lbrace{CoG}\rbrace}\bm{f} \in \mathcal{V}_{\mathcal{F}},
\end{equation}
\begin{equation}
 \| {}^{\lbrace{CoG}\rbrace}\bm{\tau} \| \le \tau_{min}\: \Rightarrow\:
  {}^{\lbrace{CoG}\rbrace}\bm{\tau} \in \mathcal{V}_{\mathcal{T}}.
\end{equation}\par
Note that our robot is controled under the premise that the roll and
pitch angles are proximate to zero. As such, we posit that the gravity
force is horizontal to the {CoG} frame, and we account for this force as
an offset when defining the guaranteed control force.\par
\begin{figure}[!t]
 \begin{center}
   \includegraphics[width=0.8\columnwidth]{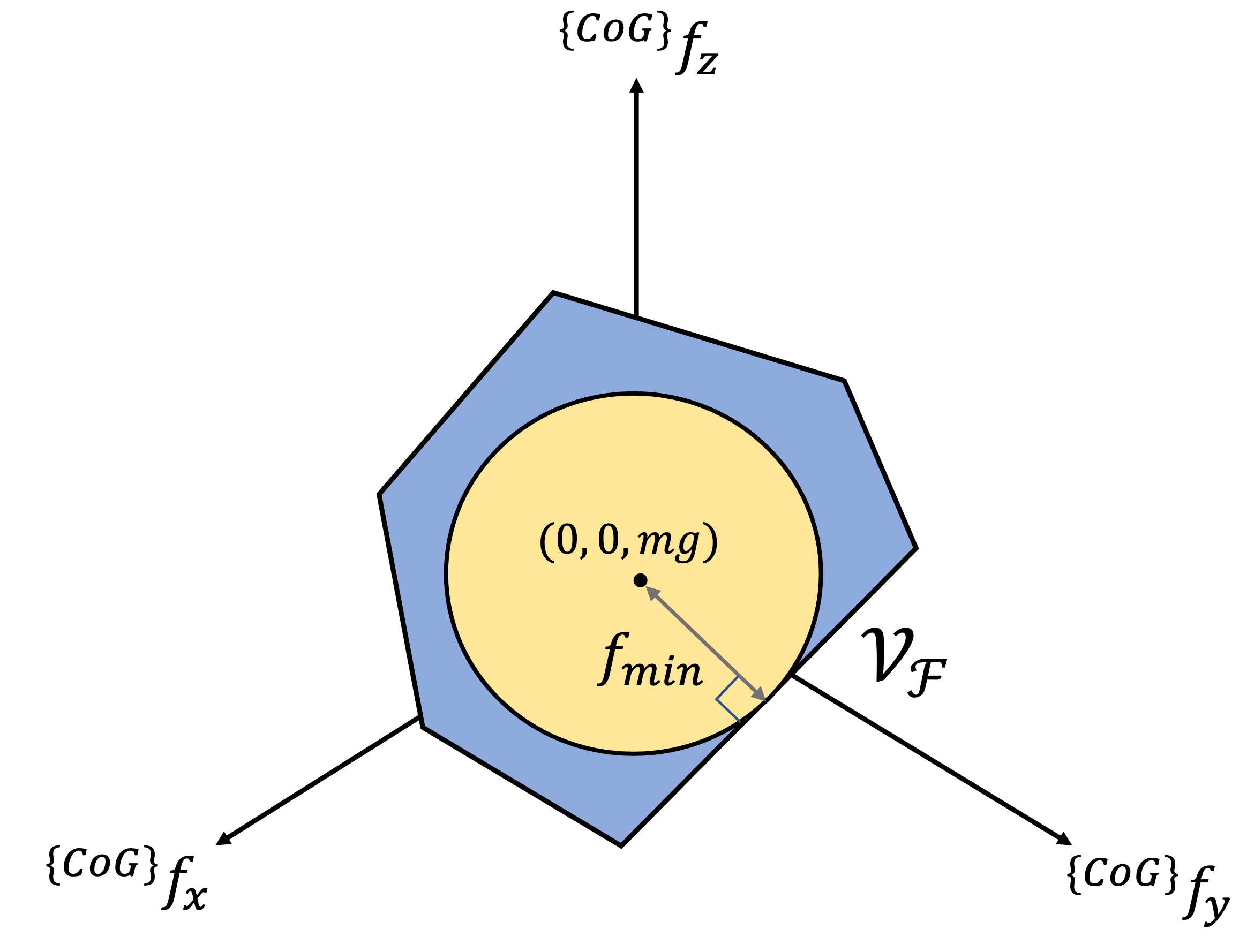}
 \end{center}
 \caption{The example of a feasible control force convex polyhedron defined in
 \eqref{equation:force_convex}. The blue region in this figure is
 $\mathcal{V}_{\mathcal{F}}$ , and ${}^{\lbrace{CoG}\rbrace}\bm{f}$ in the
 yellow region satisfies the left side of \eqref{equation:f_min}.}
   \label{figure:convex}
\end{figure}

Therefore, by maximizing these $f_{min}$ and $\tau_{min}$, the guaranteed force and torque regions
can be maximized. Thus, we initially explicate the methodology for
computing $f_{min}$.
As an example of feasible control force convex polyhedron which is depicted in
\figref{figure:convex}, $f_{min}$ is equal to the radius of the
inscribed sphere of this polyhedron, and the same is true for
torque. Thereby, $f_{min}$ is calculated by exploiting the distance $d^{f}_{ij}\left(\bm{U}\right)$, which is the length from
the origin to a plane of polyhedron  along its normal vector $\bm{h^{f}_{ij}}$.

The calculation of $d^{f}_{ij}\left(\bm{U}\right)$ can be performed as following.
\begin{equation}
 d_{i j}^{f}\left(\bm{U}\right) =\left|\sum_{k=1}^{N} \max \left(0, \lambda_{\max }\:^{t}\bm{h^{f}_{ij}}
                                               {}^{\lbrace{CoG}\rbrace}\bm{u_{k}}\right)
  -
  \:^{t}\bm{h^{f}_{ij}}m
  \boldsymbol{g}\right|,
\end{equation}
where
\begin{equation}
 \bm{h^{f}_{ij}} =
 \frac{
      {}^{\lbrace{CoG}\rbrace}\bm{u_{i}}
             \times
             {}^{\lbrace{CoG}\rbrace}\bm{u_{j}}}
      {
      \left|{}^{\lbrace{CoG}\rbrace}\bm{u_{i}}
             \times
             {}^{\lbrace{CoG}\rbrace}\bm{u_{j}}\right|
             } \:\: .
\end{equation}
Moreover, as $f_{min}$ corresponds to the radius of the inscribed sphere,
we may ascertain $f_{min}$ in the following manner:
\begin{equation}
 f_{min}=\min _{i, j} d_{i j}^{f} .
\end{equation}
\par
Note that, if $f_{min} > 0$, the robot can fly.\par
Similarly, $\bm{\tau_{min}}$ can be acquired by calculating $d_{i j}^{\tau}$
in the following manner.
\begin{equation}
 d_{i j}^{\tau} \left(\bm{U}\right)=\left|\sum_{k=1}^{N} \max \left(0, \lambda_{\max }
                                           \:{}^{t}\bm{h^{\tau}_{ij}}
                                           {}^{\lbrace{CoG}\rbrace}\bm{v_{k}}\right)\right|,
\end{equation}

\begin{equation}
 \tau_{min}=\min _{i, j} d_{i j}^{\tau},
\end{equation}
where
\begin{equation}
 \bm{h^{\tau}_{ij}} =
 \frac{
      {}^{\lbrace{CoG}\rbrace}\bm{v_{i}}
             \times
             {}^{\lbrace{CoG}\rbrace}\bm{v_{j}}}
      {
      \left|{}^{\lbrace{CoG}\rbrace}\bm{v_{i}}
             \times
             {}^{\lbrace{CoG}\rbrace}\bm{v_{j}}\right|
             } \:\: .
\end{equation}
To sum up, the objective function to be maximized in this optimization of TRADY's rotor
configuration is formulated as follows.
% \begin{equation}\label{equation:ob_func}
% \begin{aligned}
% & \underset{\bm{U}}{\text{maximize}} && w_{1}f_{min} + w_{2}\tau_{min}.\\
% \end{aligned}
  % \end{equation}
\begin{equation}\label{equation:ob_func}
 S\left(\bm{U}\right) =  w_{1}f_{min} + w_{2}\tau_{min},
\end{equation}
where $w_{1}$ and $w_{2}$ are the positive weights to balance between
force and torque.
Additionally, the constraints can be expressed as follows:
\begin{equation}\label{equation:const_force}
  \begin{aligned}
   & \bm{\ast 1}: &&
  f_{min} > 0,
  \end{aligned}
\end{equation}
\begin{equation}\label{equation:const_torque}
   \begin{aligned}
   & \bm{\ast 2}: &&
    \tau_{min} > 0.
  \end{aligned}
\end{equation}
Although our primary focus is on the assembled aircraft consisting of
two quadrotor units, the maximization of $S\left(\bm{U}\right)$ within the constraints
described in \eqref{equation:const_force} and \eqref{equation:const_torque} ensures the attainment of fully-actuated
model control in aircraft composed of an arbitrary number of units.
\subsubsection{Under-actuated Model Controllability}
\begin{figure}[!t]
 \begin{center}
   \includegraphics[width=\columnwidth]{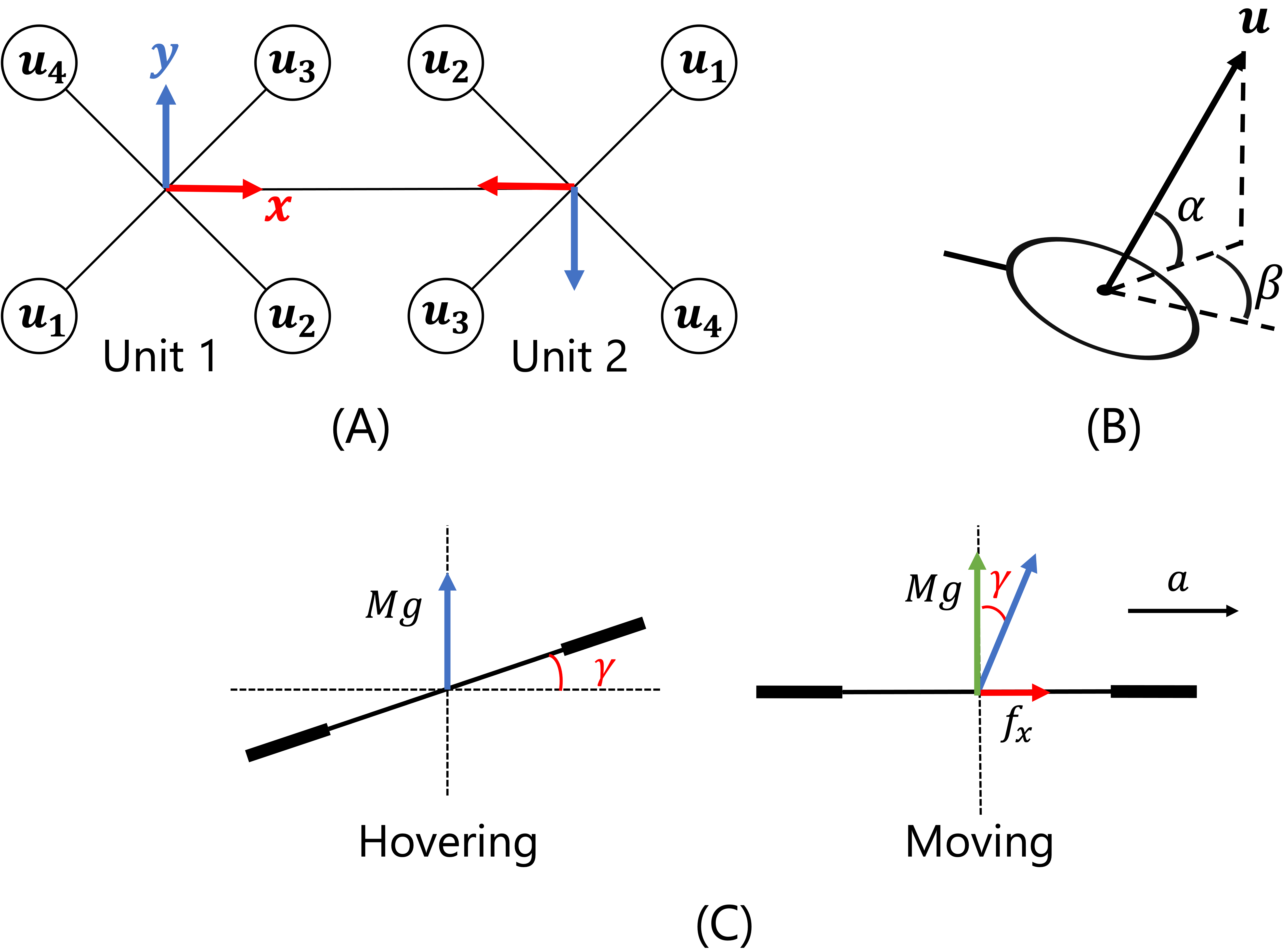}
 \end{center}
 \caption{(A)Overall configuration of TRADY in its assembled state. In
 this configuration, each docking mechanism is mounted on the same
 position of the unit.(B)Rotor direction vector $\bm{u}$ represented by
 spherical coordinate parameters $\alpha$ and $\beta$. (C)Airframe
 inclination during hovering and movement.}
   \label{figure:unit_config}
\end{figure}

Subsequently, we present the constraints for \eqref{equation:ob_func} to enable TRADY
to achieve stable flight in the unitary state. Firstly, as previously
noted in \secref{sec:introduction}, the TRADY is composed of two quadrotor units, each
possessing an identical rotor configuration. Consequently, the whole
structure can be illustrated as \figref{figure:unit_config}(A), with a
constraint expressed as follows:
\begin{equation}\label{equation:constraint_1}
 \bm{u_{1}} = \bm{u_{5}},  \bm{u_{2}} = \bm{u_{6}},  \bm{u_{3}} =
  \bm{u_{7}},  \bm{u_{4}} = \bm{u_{8}} .
\end{equation}
Additionally, as noted above, the prerequisite for stable flight of the units is that
both \eqref{equation:torquecondition2} and \eqref{equation:forcecondition2} are satisfied. Despite the presence of an
infinite number of combinations for ${\lbrace{C}\rbrace}$ and
$\bm{\lambda_{s}}$, we assume that $\bm{\lambda_{s}} =
\lambda_{s}\;{}^{t}\lbrack\begin{array}{cccc}
 1&1 &1 &1 \\
       \end{array}\rbrack$, to
ensure even distribution of workload across each rotor. Consequently, an
additional constraint can be formulated as follows.
\begin{equation}\label{equation:constraint_2}
  \begin{aligned}
   & \bm{\ast 3}: &&
  \|\lambda_{s} \bm{Q_{trans}}\; {}^{t}\lbrack\begin{array}{cccc}
 1&1 &1 &1 \\
                                             \end{array}\rbrack\;\|
                                             = mg,
   \end{aligned}
\end{equation}

\begin{equation}\label{equation:constraint_3}
   \begin{aligned}
   & \bm{\ast 4}: &&
  \lambda_{s} \bm{Q_{rot}}\; {}^{t}\lbrack\begin{array}{cccc}
 1&1 &1 &1 \\
                                             \end{array}\rbrack
                                             = \bm{0} .
  \end{aligned}
\end{equation}
\subsubsection{Flight Characteristics for Aerial Assembly/Disassembly}
Next, we consider the flight characteristics required for aerial
assembly/disassembly. Generally, quadrotors obtain translational
propulsion by inclining their airframe. However, this inclination impedes
parallel contact between two units during assembly motion. The coupling
mechanism proposed in this study has a small contact area, which renders
this non-parallel relationship a hindrance to the assembly
process.\par
However, in the instance of the TRADY unit, its target
coordinate system is ${\lbrace{C}\rbrace}$, which is inclined with respect to
${\lbrace{CoG}\rbrace}$. Consequently, the airframe inclines during hovering, and
conversely, it assumes a horizontal orientation during movement in a
specified direction. Hence, we establish a novel constraint for the optimization to regulate the tilted angle of ${\lbrace{C}\rbrace}$ in such a way
that the airframe assumes a horizontal orientation when it accelerates along
the positive x-axis direction in \figref{figure:unit_config}(B).\par
Assuming
that the inclination angle during hovering is $-\gamma$ as
shown in \figref{figure:unit_config}(C), the force exerted by the
unit in the x-direction when it is in a horizontal state can be written as
$mg\tan\gamma$. Then the desired value of $\gamma$ is
calculated as follows:
\begin{equation}
 \gamma = -\tan^{-1}\left(\frac{^{des}a}{g}\right),
\end{equation}
where $^{des}a$ is the target acceleration in the x-direction.
Consequently, we introduce the following constraint condition for the
rotation matrix $\bm{R_{C}}$, to ensure that ${\lbrace{C}\rbrace}$ tilts by $-\gamma$ around the
y-axis:
\begin{equation}
  \begin{aligned}
   & \bm{\ast 5}: &&
   \bm{R_{c}} = \left[
\begin{array}{ccc}
 \cos\left(-\gamma\right)&0 &\sin\left(-\gamma\right) \\
 0&1 &0 \\
 -\sin\left(-\gamma\right)&0 &\cos\left(-\gamma\right) \\
\end{array}
\right] .
  \end{aligned}
\end{equation}\par
To sum up, the optimization problem for the rotor configuration can be
summarized as follows:
\begin{equation}\label{equation:prob_state}
\begin{aligned}
 & \underset{\bm{U}}{\text{maximize}} && S\left(\bm{U}\right)\\
 & \text{subject to} && \bm{\ast 1} \sim \bm{\ast 5}
\end{aligned}
  \end{equation}
\subsubsection{Solver for Optimization Problem}
Finally, in order to solve the designed optimization problem, it is
necessary to choose an algorithm, but it is not always guaranteed that a
solution that satisfies all the set constraints exists. Therefore, we
use the  global optimization algorithm ISRES \cite{Isres} that can find
the closest possible solution even if a perfect solution is not found. The optimized $\bm{U}$ obtained as a result of solving the
optimization problem using the parameters shown in \tabref{table:op_param} and the
guaranteed minimum force and torque in the unitary and assembly states
are presented in \tabref{table:op_result}. Here, each $\bm{u}$ is represented by spherical
coordinate parameters $\alpha$ and $\beta$ as shown in
\figref{figure:unit_config}(B). Furthermore, the outcomes are rounded off to two significant digits. The obtained outcome reveals that the
optimized rotor angles exhibit a symmetrical pattern. This occurrence
can be attributed to the constraints established in
\eqref{equation:constraint_2} and \eqref{equation:constraint_3}, as
these equations embody force and torque offsets.
In addition to this, it can be seen that both the minimum guaranteed force and torque become more than twice as
high in the assembly state when using fully-actuated model compared to the
unitary state.
\renewcommand{\arraystretch}{1.5}
\begin{table}[!t]
 \caption{Parameters for optimization.}
 \label{table:op_param}
\centering
\begin{tabular}{|c|c|}
 \hline
 Parameter &Value \\
 \hline
 Mass of a unit   &  \SI{1.1}{kg}\\
 Coordinates of rotors  &  $\left(\pm\:0.12\:\text{m},\pm\:0.12\:\text{m}\right)$\\
 Maximum thrust of rotor  & \SI{7}{N}\\
 Coefficient of counter-torque  & \SI{-0.011}{\per\meter} \\
 $^{des}a$& \SI{1}{m/s^2}\\
 \hline
\end{tabular}
\end{table}
\renewcommand{\arraystretch}{1.5}
\begin{table}[!t]
 \caption{Result optimization.}
 \label{table:op_result}
\centering
\begin{tabular}{|c|c|}
 \hline
 Parameter &Value \\
 \hline
 $\lbrack\alpha_{1},\; \beta_{1}\rbrack$ & \lbrack 0.45\:\text{rad},\; 0.73\:\text{rad}\rbrack\\
 $\lbrack\alpha_{2},\; \beta_{2}\rbrack$ & \lbrack 0.52\:\text{rad},\; -2.1\:\text{rad}\rbrack\\
 $\lbrack\alpha_{3},\; \beta_{3}\rbrack$ & \lbrack 0.52\:\text{rad},\; 2.1\:\text{rad}\rbrack\\
 $\lbrack\alpha_{4},\; \beta_{4}\rbrack$ & \lbrack 0.45\:\text{rad},\; -0.73\:\text{rad}\rbrack\\
 \large$f_{min,\:unit}$ & \SI{3.1}{N}\\
 \large$\tau_{min,\:unit}$ & \SI{0.64}{Nm}\\
 \large$f_{min,\:assem} $ & \SI{7.5}{N}\\
 \large$\tau_{min,\:assem}$ & \SI{2.8}{Nm}\\
 \hline
\end{tabular}
\end{table}
\renewcommand{\arraystretch}{1.0}

\section{Control}\label{sec:control}
Firstly, we outline the methodology for flight control, which
incorporates fully-actuated model control for the unitary state and
fully-actuated model control for the assembly state. Next, we propose a system switching
strategy to be executed during the docking or undocking process.
\subsection{Fully-actuated Flight Control}
\begin{figure*}[!t]
 \begin{center}
   \includegraphics[width=1.4\columnwidth]{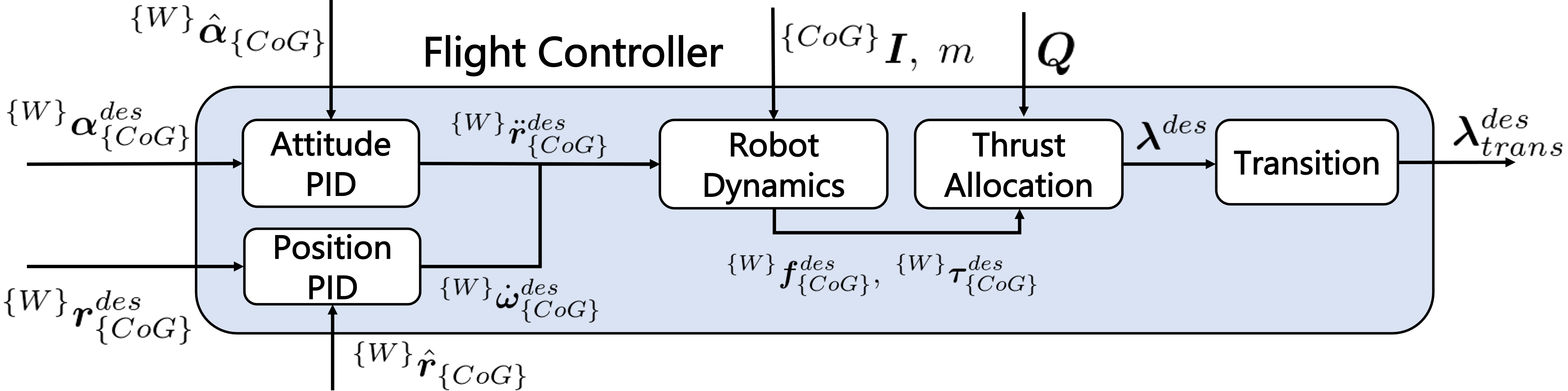}
 \end{center}
 \caption{Block diagram of fully-actuated controller. The blue rectangle represents the processing flow of the
 controller, while the incoming arrows from other components shown in
 \figref{figure:whole_system} denote parameter transmission. The ``Robot Dynamics'' in
 this diagram refers to the Newton-Euler equations described in
 \eqref{equation:translation} and \eqref{equation:rotation}, and ``Thrust allocation'' is the computation based on
 the pseudoinverse matrix, as described in \eqref{equation:Qinv}. Additionally, ``Transition'' is the state transition process defined by equation 3.
 }
   \label{figure:a_fc}
\end{figure*}
\subsubsection{Position Control}
For position control, a general PID controller is adopted. Thus, from \eqref{equation:force},
\eqref{equation:torque}, the desired force and torque can be derived as
following:
\begin{equation}\label{equation:des_f}
 \bm{f}^{des} = m\;{}^{\lbrace W\rbrace}\bm{R}_{\lbrace CoG \rbrace}^{-1}\left(\bm{K_{p}}\bm{e_{r}} +\bm{K_{i}}\int\bm{e_{r}}dt
                       +\bm{K_{d}}\dot{\bm{e_{r}}}\right),
\end{equation}\\
where $\bm{e_{r}}$ is a position error defined as $\bm{e_{r}} =
{}^{\lbrace W \rbrace}\bm{r}^{des}_{\lbrace CoG \rbrace} - {}^{\lbrace W \rbrace}\bm{r}_{\lbrace CoG \rbrace}$, and
$\bm{K_{p}}$, $\bm{K_{i}}$, $\bm{K_{d}}$ are gains for
controller. Additionally, ${\lbrace C \rbrace}$ is used instead of
${\lbrace CoG \rbrace}$ in the unitary state.\par
In the assembly state, as shown in \figref{figure:a_fc}, the obtained $\bm{f}^{des}$ from
\eqref{equation:des_f} is allocated to the desired thrust by $\bm{Q_{trans}}$.
\subsubsection{Attitude Control}
Next, we explain attitude control. In the case of a fully-actuated
model, it is possible to apply traditional PID control for attitude
control. Here, by using \eqref{equation:torque}, the desired torque can
be obtained as follows:
\begin{equation}
 \begin{split}
 \bm{\tau}^{des} =\;{}^{\lbrace{CoG}\rbrace}\bm{I}\left(\bm{K_{p}}\bm{e_{\alpha}} +\bm{K_{i}}\int\bm{e_{\alpha}}dt
  +\;\bm{K_{d}}\dot{\bm{e_{\alpha}}}\right) \\
   +
  {}^{\lbrace{CoG}\rbrace}\bm{\omega}\times\;{}^{\lbrace{CoG}\rbrace}\bm{I}\;{}^{\lbrace{CoG}\rbrace}\bm{\omega},
   \end{split}
\end{equation}
where $\bm{e_{\alpha}}$ is an attitude error defined as $\bm{e_{\alpha}} = {}^{\lbrace W
\rbrace}\bm{\alpha}^{des}_{\lbrace{CoG}\rbrace} - {}^{\lbrace W \rbrace}\bm{\alpha}_{\lbrace{CoG}\rbrace}$ and
$\bm{K_{p}}$, $\bm{K_{i}}$, $\bm{K_{d}}$ are gains for controller.
Then, the desired torque $\bm{\tau}^{des}$ obtained from attitude control is allocated into the
desired thrust for each rotor using the allocation matrix $\bm{Q_{rot}}$
as shown in \figref{figure:a_fc}.
\subsection{Under-actuated Flight Control}
\begin{figure*}[!t]
 \begin{center}
   \includegraphics[width=1.4\columnwidth]{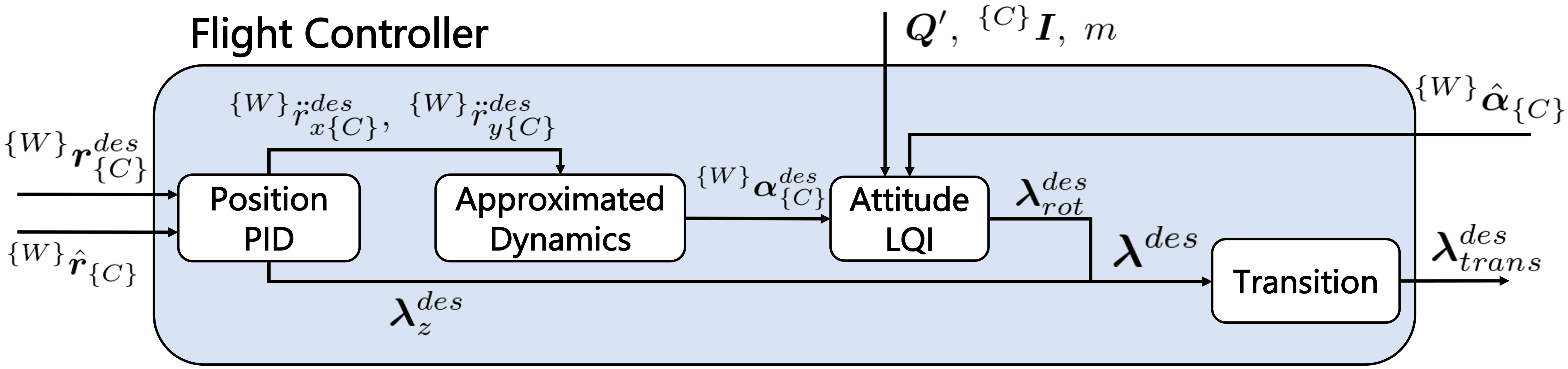}
 \end{center}
 \caption{Block diagram of under-actuated controller. ``Approximated Dynamics'' in this diagram represents the operation
 of converting the target acceleration to the target attitude through
 the approximation of the horizontal state described by \eqref{equation:theta} and
 \eqref{equation:phi}. In addition, $\bm{\lambda_{z}}^{des}$ is actually the output of ``Position PID'' converted to thrust by \eqref{equation:lambda_z}.
 }
   \label{figure:u_fc}
\end{figure*}

\subsubsection{Position Control}
The position control for the underactuated model is developed using the
methodology presented in \cite{Lee2010}, which involves determining the
target force via conventional PID feedback loop and subsequently converting
it to target roll and pitch angles. The target forces $f^{des}_x$ and $f^{des}_y$ can be
obtained using \eqref{equation:des_f}, followed by determining the
target roll $\theta$ and
pitch $\phi$ as outlined below:
\begin{equation}\label{equation:theta}
{}^{\lbrace W\rbrace}\theta^{des}_{\lbrace C \rbrace} =
 atan^{-1}\left(-f_{y}, \sqrt{f^{2}_{x}+f^{2}_{z}}\right),
\end{equation}
\begin{equation}
{}^{\lbrace W\rbrace}\phi^{des}_{\lbrace C \rbrace} =
 atan^{-1}\left(f^{2}_{x}, f^{2}_{z}\right).
\end{equation}\label{equation:phi}
These target angles are achieved by attitude controller.\par
Additionally, regarding the position control in z-direction, collctive thrust force $f^{des}_{z}$ is calculated in
the following manner:
\begin{equation}
 f^{des}_{z} = ^{t}\left({}^{\lbrace W\rbrace}\bm{R}_{\lbrace C \rbrace}\bm{b}\right)\bm{f}^{des},
\end{equation}
where $\bm{b}$ is a unit vector $^{t}\left[\begin{array}{ccc}
                                  0&0 &1 \\
                                       \end{array}\right]$.\par
Then, using $\bm{\lambda_{s}}$ defined in
\eqref{equation:force_condition} and \eqref{equation:torque_condition},
the target thrust for z-directional control is calculated as follows:
\begin{equation}\label{equation:lambda_z}
 \bm{\lambda}^{des}_{z} = \frac{f^{des}_{z}}{mg}\bm{\lambda}_{s}.
\end{equation}
\subsubsection{Attitude Control}
In attitude control, to suppresse the uncontrollable horizontal forces
due to the tilted rotors, we adopt LQI control\cite{LQI} which is a type of optimal control that derives control inputs that minimize the cost function. Therefore, by designing the cost function appropriately, various requirements can be met in addition to convergence speed. The state equation of posture control is described as follows:
\begin{align}\label{equation:state_func}
& \bm{\dot{x}} = \bm{A}\bm{x}+
  \bm{B}\bm{\lambda}+\bm{D}\left(\;{}^{\lbrace C \rbrace}\bm{I}^{-1}\;{}^{\lbrace C\rbrace}\bm{\omega}\times\;{}^{\lbrace C \rbrace}\bm{I}\;{}^{\lbrace C\rbrace}\bm{\omega}
 \right), \notag \\
& \bm{y} = \bm{C}\bm{x},
\end{align}
where
\begin{align}
& \bm{x}  =\; ^{t}\left[\begin{array}{ccccccccc}
            e_{x} & \dot{e_{x}} & e_{y}  & \dot{e_{y}} & e_{z}  & \dot{e_{z}} &\int{e_{x}} &
             \int{e_{y}} & \int{e_z}\end{array}\right], \notag\\[+10pt]
& \bm{e} = \bm{\alpha}^{des} - \bm{\alpha}, \notag \\[+10pt]
& \bm{\dot{e}} = \dot{\bm{\alpha}}^{des} - \dot{\bm{\alpha}}, \notag \\[+10pt]
& \bm{B} =\; ^{t}\left[\begin{array}{ccccccc}
           \bm{0_{4\times1}}&\bm{B_{1}}&\bm{0_{4\times1}} &\bm{B_{2}} &\bm{0_{4\times1}} &\bm{B_{3}} &\bm{0_{4\times3}}\\
                  \end{array}\right], \notag\\[+10pt]
&  \left[
 \begin{array}{ccc}
 \bm{B_{1}}&\bm{B_{2}} &\bm{B_{3}} \\
 \end{array}
 \right]
  = \;^{t}\left({}^{\lbrace C \rbrace}\bm{I}^{-1}\bm{Q_{rot}}'\right).
\end{align}
Note that the target roll and pitch angle is ${}^{\lbrace
W\rbrace}\theta_{\lbrace C \rbrace}$, ${}^{\lbrace
W\rbrace}\phi_{\lbrace C \rbrace}$ obtained from \eqref{equation:theta} and \eqref{equation:phi}.
In regards to the cost function, this study designs a function with the
objective of improving convergence, suppressing the control input, and suppressing translational forces. In this case, the cost function is given as follows:
\begin{equation}\label{equation:cost_func}
J = \int_{0}^{\infty} \left(\;^{t}\bm{x}\bm{M}\bm{x} + \;^{t}\bm{\lambda}\bm{N}\bm{\lambda}\right)dt,
\end{equation}
where $\bm{M}$ and $\bm{N}$ are diagonal weight matrices. The first term
in \eqref{equation:cost_func} corresponds to the control output's norm,
and minimizing this norm can enhance the convergence. Moreover,
concerning the second term in \eqref{equation:cost_func}, the method of defining $\bm{N}$ as follows has been proposed in \cite{singularity}:
\begin{equation}\label{equation:weight}
 \bm{N} = \bm{W_{1}} + \bm{Q_{trans}}^{T}\;'\bm{W_{2}}\bm{Q_{trans}}',
\end{equation}
where $\bm{W_{1}}$ and $\bm{W_{2}}$ are also diagonal weight
matrices. Then, the first term of \eqref{equation:weight} creates a
quadratic form $^{t}\bm{\lambda}\bm{N}\bm{\lambda}$ when substituted into \eqref{equation:cost_func}, corresponding to
the norm of control input. Therefore, by minimizing this term, the
control input can be suppressed. Furthermore, when the second term is
substituted into \eqref{equation:cost_func}, the following equation is derived.

\begin{equation}\label{equation:force_norm}
 ^{t}\bm{\lambda}\;^{t}\bm{Q_{trans}}'\bm{Q_{trans}}'\bm{\lambda}
  = \;^{t}\bm{f}\bm{f} = \|\bm{f}\|^{2}.
\end{equation}
As seen from \eqref{equation:force_norm}, the norm of translational
force is represented by the second term of \eqref{equation:weight}. Therefore, by employing the control input $\bm{\lambda}$ that minimizes the cost function
defined in \eqref{equation:cost_func}, stable attitude control of a unit
equipped with tilted rotors can be achieved.
By solving the algebraic Riccati equations derived from
\eqref{equation:state_func} and \eqref{equation:cost_func}, we obtain
the feedback gain $\bm{K}$. Therefore, the desired
$\bm{\lambda}$ is calculated as follows:
\begin{equation}
 \bm{\lambda}^{des}_{rot} = \bm{K}\bm{x}+
  {}^{\#}\bm{Q_{rot}}'\;{}^{\lbrace
  C\rbrace}\bm{\omega}\times\;{}^{\lbrace C\rbrace}\bm{I}\;{}^{\lbrace
  C\rbrace}\bm{\omega} .
 \end{equation}\par
Finally, in combination with the $z$ axis position control, the final
output in the unitary state is ultimately calculated as follows:
\begin{equation}
 \bm{\lambda}^{des} = \bm{\lambda}^{des}_{z} + \bm{\lambda}^{des}_{rot} .
\end{equation}

\subsection{System Switching}\label{subsec:transition}
\begin{figure*}[!t]
 \begin{center}
   \includegraphics[width=1.6\columnwidth]{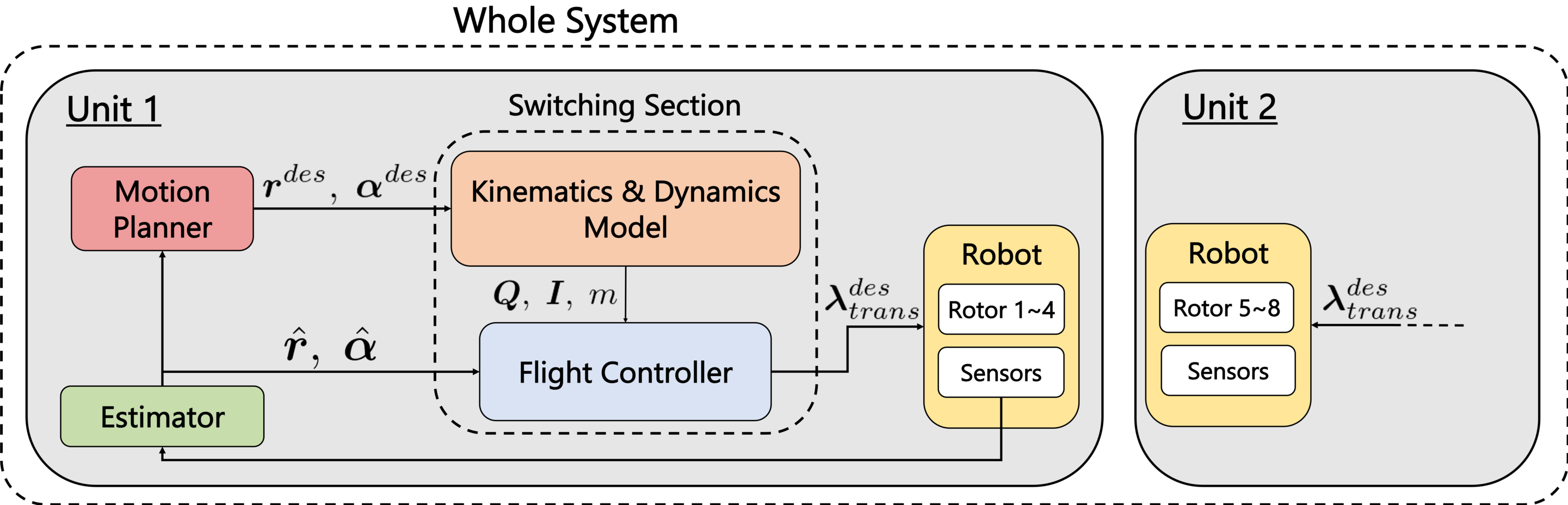}
 \end{center}
 \caption{Overall system diagram of the control system. The ``Estimator''
 in the figure processes various sensor values using a Kalman filter to
 perform self-state estimation. Additionally, the ``Motion Planner'' sends
 position and attitude commands for assembling/disassembling actions and
 aerial task realization. The ``Switching Section'' is a partition that
 switches between the assembly state and the unitary state, which includes
 the ``Kinematics \& Dynamics Model'' and ``Flight Controller''. The
 ``Kinematics \& Dynamics Model'' outputs the inertia values of the
 aircraft and $\bm{Q}$ in the case of under-actuated or fully-actuated model, while the ``Flight
 Controller'' is shown in \figref{figure:a_fc} and
 \figref{figure:u_fc}. Even in the assembly state, the
 units do not communicate with each other and independently control
 their respective rotors.
 }
   \label{figure:whole_system}
\end{figure*}
We present a methodology for system switching that is executed
during assembling and disassembling actions.
\subsubsection{Overall Control System}
To begin with, the overall
control system is designed, as depicted in \figref{figure:whole_system}. The ``Motion
Planner'', ``Estimator'', and ``Robot'' sections adopt common processes in
both the assembly state and the unitary state. Conversely, the
``Switching Section'' serves as a partition that alternates between the
assembly and unitary states. During the assembly state, the entire
system is controlled by a distributed control system, while each unit
controls its respective rotors based on a common robot model without
communicating with one another.
Therefore, the overall structure of the system remains unchanged between
the assembly state and the unitary state, with only the contents of the
``Switching Section'' being switched.
\subsubsection{Transition Process}
Next, we explain the transition process that should be performed when
switching systems. The problem when switching the system is the model
error related to the inertia values and rotor performance of the
robot. In normal flight control of TRADY, model errors are compensated
for by the integral term of PID and LQI. In addition, there is no offset
thrust based on default values in the gravity direction, and it is all
compensated for by the integral term. Model error compensation by
integral value is gradually performed while gradually increasing the
rotor thrusts during takeoff, so it is generally completed until
hovering.\par
As an example, we consider the system switching during the assembly
action. When switching the system, most control values are reset because
the controller is switched. However, as mentioned earlier, gravity
compensation is performed by the integral term, so only the integral
terms are carried over in order to continue the flight. However, since
this integral term compensates for the model error of the robot in the
unitary state, the model error in the assembly state is not compensated
for. Model errors that suddenly occur during flight, unlike during
takeoff, cause discrete increases or decreases in rotor thrusts, which
have a significant negative effect on control stability.\par

To address this issue, a method of gradually transitioning rotor thrusts
while compensating for the model error of the assembly state model was
developed. Although it is impossible to predict each rotor thrust after
model error compensation, the total weight of the system does not change
before and after system switching, thereby the total thrust force acting in
the gravity direction after model error compensation can be said to be
the same as in the unitary state. Furthermore, since among each rotor thrust,
the proportion that works in the direction of gravity does not change,
consequently, the total thrust of all rotors can be said to be equal before and
after switching. Therefore, the influence of sudden model errors can be
suppressed by scaling the rotor thrusts output from the controller based
on the total thrust in the unitary state. However, it is necessary to
reflect the influence of the control output in the assembly state,
thereby the control output is scaled using the value obtained by
superimposing the total thrust in these two states.\par
 Here, the total thrust in the unitary state is obtained using the total thrust
immediately before the system switch, and the total thrust in the
assembly state is obtained using the real-time control output. Then, rotor thrusts during the transition is scaled as follows:
\begin{equation}\label{equation:trans_lambda}
 \bm{\lambda^{des}_{trans}} =
  \frac{
  S_{trans}
  }{
  S_{assem}
  }
  \bm{\lambda^{des}_{assem}},
\end{equation}
\begin{equation}\label{equation:trans_sum}
 S_{trans} =
  W\left(t\right)S_{assem}+
  \left(1- W\left(t\right)\right) S_{unit},
\end{equation}
where $S_{assem}$ is the total thrust in the assembly state defined as $S_{assem} =
\sum\limits_{i=1}^{n} \|\bm{\lambda}^{des}_{assem,\;i}\|$ and $S_{unit}$ is that in
the unitary state. Note that $S_{assem}$ is a variable but $S_{unit}$ is a constant because $S_{unit}$ competed
by the thrust at the moment of switching. Additionally, $W\left(t\right)$ is the weight
function used in the superimposing and it is desirable for it to
smoothly converge to 1. Therefore, $W\left(t\right)$ is defined as follows:
\begin{equation}\label{equation:weight_func}
 W\left(t\right) = 1-\frac{1}{at+1},
\end{equation}
where $a$ is a constant and $t$ is a variable representing time. Then, $W\left(t\right)$ satisfies follows:
\begin{align}
& W\left(0\right) = 0, \notag \\
& \lim_{t \to \infty}W\left(t\right) \to 1 .
\end{align}
Therefore, in \eqref{equation:trans_sum}, initially the influence of $S_{unit}$ is
dominant, but gradually $S_{assem}$ becomes dominant and ultimately $\bm{\lambda_{trans}}$ becomes
indistinguishable from the normal output of the assembly state controller. Thus, as a result, even after the system switch, the rotor thrust
transitions smoothly. The variable $a$ in \eqref{equation:weight_func} determines the
convergence rate of the function, and in this work, it was empirically
set to $a = 0.9$.
In this case, $W\left(t\right)$ attains a value of 0.99 when $t = 120$. During actual
implementation, $W\left(t\right)$ is refreshed at each control cycle with a frequency of
\SI{0.025}{s}. Consequently, $W\left(t\right)$ achieves 0.99 after approximately \SI{3}{s}
of real time have elapsed.\par
To sum up, by utilizing the aforementioned methodology, it is possible to
address the discrete thrust changes caused by model errors during system
switching. This transition process occurs outside the controller, as
illustrated in \figref{figure:a_fc} and \figref{figure:u_fc}, and can be regarded as a kind of
disturbance to the feedback loop. However, if this disturbance itself
converges sufficiently quickly, the overall system stability is
ultimately dependent on the stability of controllers. In the case of our
method, the tuning of $a$ in \eqref{equation:weight_func} ensures the
convergence rate of the transition process, while the stability of the
PID controller is achieved through the tuning of its gains, and the
stability of the LQI controller is guaranteed by
\eqref{equation:cost_func}. Therefore, the proposed transition process
has no impact on the stability of the feedback loop.

\section{Motion Strategy}\label{sec:motion_strategy}
In this section, we introduce the motion strategy for executing
assembly/disassembly motion in mid-air. Initially, we propose the
overarching strategy, followed by an specific explication of each constituent.
\subsection{Overall Strategy}
\begin{figure}[!t]
 \begin{center}
   \includegraphics[width=0.9\columnwidth]{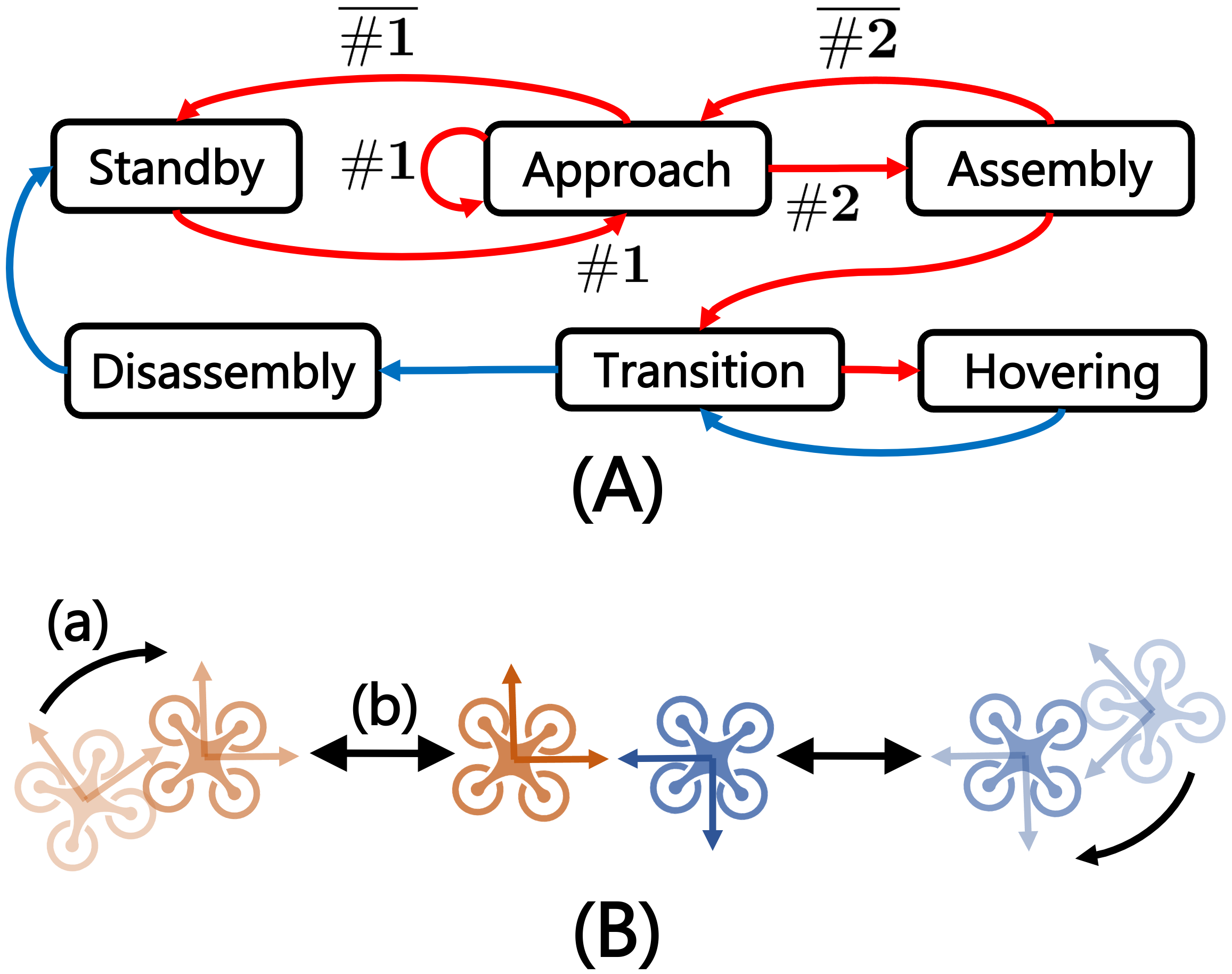}
 \end{center}
 \caption{(A)The Finite State Machine for assembly/disassembly motion:
 Each block represents a corresponding state.The red arrows represent
 the transitions between the states for assembly, while the blue ones
 represent that for disassembly. Note that $\overline{\bm{\#}}$
 represents the negation of $\bm{\#}$. (B)The procedure of assembly
 motion based on proposed strategy.  Arrow (a) symbolizes the transition
 into the ``Standby'', whereas arrow (b) signifies the iterative
 transitions into the ``Approach'' and ``Standby'' states.}
   \label{figure:state_machine}
\end{figure}

As explained in Section. \ref{subsec:docking}, in aerial assembly/disassembly motions,
the two units come into extreme proximity, causing unstable flight. This
makes high-precision position control difficult. Therefore, a motion strategy is required that ensures the certainty of the motion under
multiple conditions, and executes it only if these conditions are
met. Then, in this study, we divide the target action into several
stages and represent them as an Finite State Machine (FSM) as shown in \figref{figure:state_machine}.
In this FSM, conditions are set to transition between states, and when two units fulfill them, the motion transitions to the next state.  By repeating this process, it becomes
possible to autonomously perform aerial assembly/disassembly motion while avoiding
hazardous conditions.
\subsection{Assembly Motion}
Now, we explain each state in the FSM, focusing on the assembly motion.
\subsubsection{Standby State}
This state is the initial
state of the FSM, where both units are guided towards their respective standby
positions and yaw angles.
The female unit maintains its position and yaw angle at the start of
assembly, while the male unit is guided to a position and yaw angle
$\psi$ that
satisfies the following relationship:
\begin{align}\label{equation:standby}
 &^{\lbrace F \rbrace}\bm{r}_{\lbrace M \rbrace} =
  \;^{t}\left[
   \begin{array}{ccc}
    d^{st}&0&0
   \end{array}
 \right] , \notag\\
 &^{\lbrace F \rbrace}\psi_{\lbrace M \rbrace} = -\pi
\end{align}
where${\lbrace F \rbrace}$ represents the ${\lbrace C \rbrace}$ frame
of the female unit, while ${\lbrace M \rbrace}$ represent that of the
male unit. Additionally, $d^{st}$ is the desired distance between two
units' CoG. Note that the direction of each axis is the same as that shown in
\figref{figure:unit_config}.\par
Here, we define the following condition $\bm{\#1}$ with respect to the
errors in $r_{y}$, $r_{z}$, and $\psi$:
\begin{equation}\label{equation:hash_1}
 \begin{aligned}
  & \bm{\# 1}: &&
    \left(
   \begin{array}{c}
    ^{\lbrace F \rbrace}r_{y\lbrace M \rbrace}\\^{\lbrace F
     \rbrace}r_{z\lbrace M \rbrace} \\ ^{\lbrace F \rbrace}\psi_{\lbrace M \rbrace}\\
   \end{array}
  \right)
  \le
  \left(
   \begin{array}{c}
    e^{\# 1}_{y}\\e^{\# 1}_{z}\\e^{\# 1}_{\psi}\\
   \end{array}
   \right)
 \end{aligned}
\end{equation}
where $e^{\# 1}_{y}$,$e^{\# 1}_{z}$,$e^{\# 1}_{\psi}$ are tolerable
errors. Condition $\bm{\# 1}$ is used as a criterion to determine if the
positional relationship between the two units is safe. If the
relationship between the female unit and the male unit satisfies $\bm{\#
1}$, the state transitions to the ``Approach'' state.
\subsubsection{Approach State}
At this state, the male unit moves towards the female unit, while the
female unit maintains its position and angle. During the approach phase,
if condition $\bm{\#
1}$ is no longer met, the approach motion is interrupted and
the system transitions to the ``Standby'' state. By virtue of this process, it becomes possible to instantaneously recover from a hazardous positional state between the two units and continue with the assembly motion unimpeded.\par
Here, we define the following condition $\bm{\# 2}$ with respect to the
errors in $r_{x}$, $r_{y}$, $r_{z}$, and $\psi$:
\begin{equation}\label{equation:hash_2}
 \begin{aligned}
  & \bm{\# 2}: &&
    \left(
   \begin{array}{c}
    ^{\lbrace F \rbrace}r_{x\lbrace M \rbrace}\\^{\lbrace F \rbrace}r_{y\lbrace M \rbrace}\\^{\lbrace F
     \rbrace}r_{z\lbrace M \rbrace} \\ ^{\lbrace F \rbrace}\psi_{\lbrace M \rbrace}\\
   \end{array}
  \right)
  \le
  \left(
   \begin{array}{c}
    e^{\# 2}_{x}\\e^{\# 2}_{y}\\e^{\# 2}_{z}\\e^{\# 2}_{\psi}\\
   \end{array}
   \right)
 \end{aligned}
\end{equation}
where $e^{\# 2}_{x}$,$e^{\# 2}_{y}$,$e^{\# 2}_{z}$,$e^{\# 1}_{\psi}$ are
tolerable errors. Condition $\bm{\# 2}$ is used as a criterion to determine if
the two units are in a position where they can docking. If the condition $\bm{\# 2}$ is satisfied, the state transitions
to the ``Assembly'' state.
\subsubsection{Assembly State}
In this state, initially, the docking mechanism is activated by
turning on the magnet and inserting the pegs, and subsequently, the
system is transitioned into the assembly mode. During the aforementioned movements, if condition $\bm{\# 2}$ is no longer met, the movement is halted and the state transitions to the ``Approach'' state.
\subsubsection{Transition State}
In this state, the process of transition presented in
Section. \ref{subsec:transition} is executed.
\subsection{Disassembly Motion}
In the case of aerial disassembly, the system transitions from ``Hovering'' to
``Transition,'' and then proceeds to ``Disassembly.'' During ``Disassembly,''
the coupling mechanism is released, allowing the two units to
separate. Unlike in assembly, positional control of the bodies is not
critical during disassembly, and therefore, there are no conditions that
must be satisfied for the state transitions.

\section{Experiment}\label{sec:experiment}
In this section, we first describe the robot platform utilized in each
experiment. Next, we present the experimental results, which include the
evaluation of flight stability in each state, aearil assembly and disassembly motion based on proposed control method and motion strategy, and object manipulation in the assembly state.
\subsection{Robot Platform}
\subsubsection{Hardware}
\begin{figure}[tb]
 \begin{center}
   \includegraphics[width=\columnwidth]{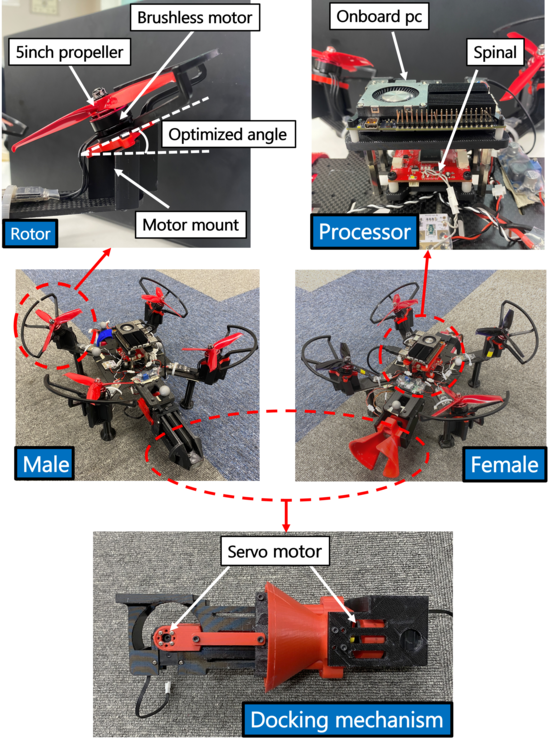}
 \end{center}
 \caption{Hardware configuration of proposed robot TRADY.}
   \label{figure:hardware_config}
\end{figure}
Based on the design proposed in \secref{sec:mech_design} and \secref{sec:rotor_config}, we introduce the
hardware configuration illustrated in \figref{figure:hardware_config}. The frame of the body is made of a \SI{5}{mm}
CFRP plate, while the other parts are mainly made of PLA material. In
determining the size of the entire body, we first determined the size of
the docking mechanism based on the rigidity required for the coupling
portion and then selected the minimum body size that can accommodate the
docking mechanism.\par

The rotors are composed of 5 inch propellers (GEMFAN Hulkie 5055S-3) and
brushless motors (ARRIS S2205), driven by ESC (T-Motor F45A). Each rotor
is tilted based on the optimized $\bm{U}$ in \secref{sec:rotor_config}. With this
configuration, each rotor exerts a thrust of approximately \SI{1}{N}-\SI{8}{N} at a
voltage of \SI{15}{V}.
\par

Furthermore, an on-board PC (Khadas VIM4) and a flight controller called
Spinal, which uses an STM32 microcontroller, are installed on the
body. The housing of the docking mechanism is made of PLA, and a servo
motor (KONDO KRS-3302) is installed on both the male and female
sides. In addition, Arduino nano is used to control the servo motor.\par

In addition, a \SI{2200}{mAh} four-cell LiPo battery is used as the power
source. However, in some experiments, power is supplied from a
bground-based stabilizing power source instead of a battery to enable
emergency stop.\par

Finally, \tabref{table:hard_prop} shows the characteristics of the robot.
\begin{table}[tb]
  \renewcommand{\arraystretch}{1.3}
  \caption{main parameters of robot}
  \centering
 \begin{tabular}{|c|c|}
  \hline
  Parameter&Value \\
  \hline
  Mass of main body& \SI{0.94}{kg}\\
  Size of main body& \SI{0.24}{m} $\times$\SI{0.24}{m}\\
  Mass of docking mechanism (Male)&\SI{0.16}{kg}\\
  Mass of docking mechanism (Female)&\SI{0.16}{kg}\\
  Continuous flight time&\SI{25}{min}\\
  \hline
 \end{tabular}
 \label{table:hard_prop}
\end{table}
\renewcommand{\arraystretch}{1.0}
\subsubsection{Software}
\begin{figure}[tb]
 \begin{center}
   \includegraphics[width=\columnwidth]{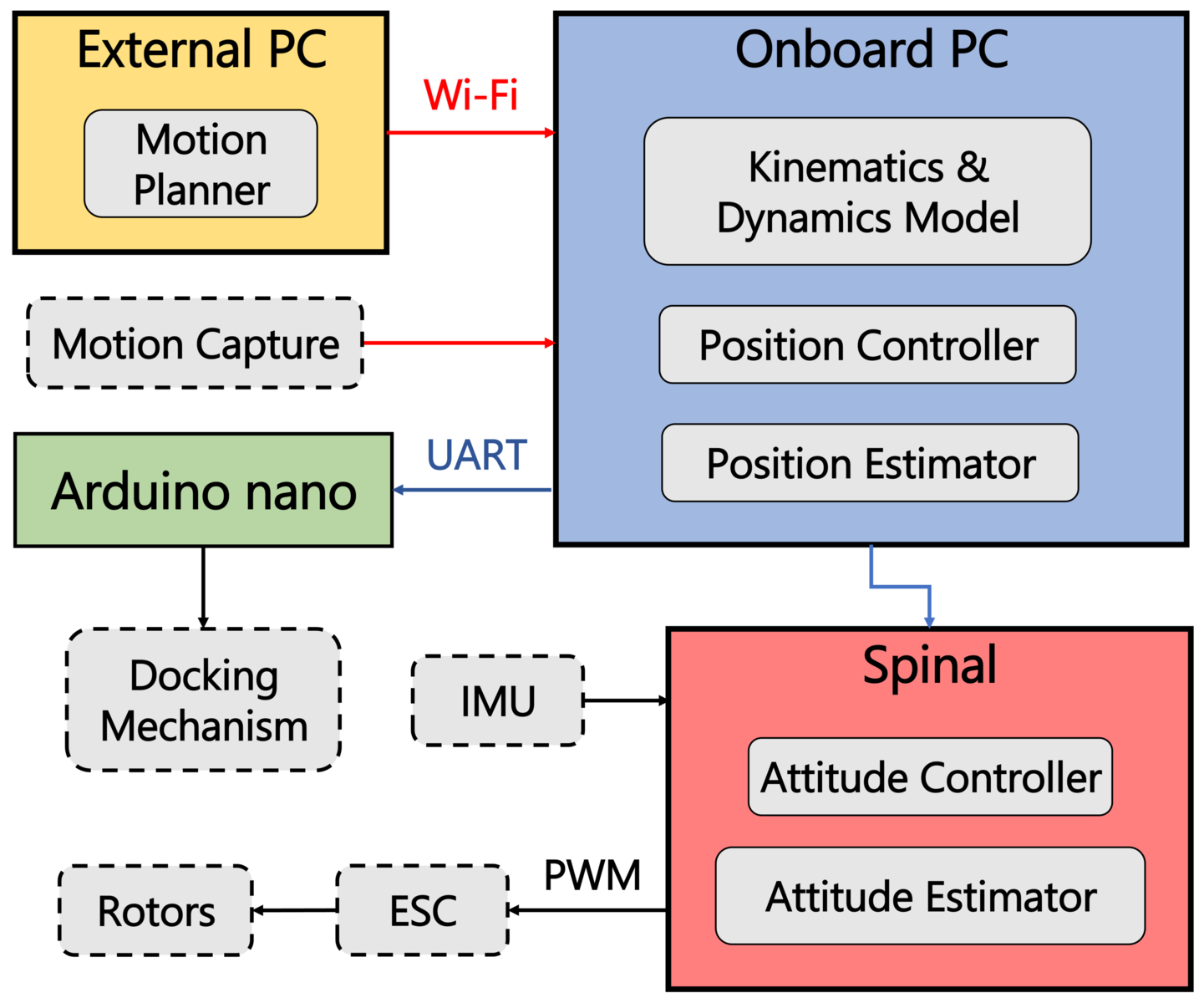}
 \end{center}
 \caption{Communication system configuration of proposed robot TRADY.}
   \label{figure:software_config}
\end{figure}
Next, we introduce the communication system for proposed robot TRADY as
shwon in \figref{figure:software_config}. Firstly, the external PC executes
motion planning including the strategy for the assembly/disassembly
motion presented in \secref{sec:motion_strategy}, and it outputs
position attitude commands.
Subsequently, these commands are transmitted to the onboard PC through a
Wi-Fi network. Furthermore, the external motion capture system transmits
the positional state of the robot to the onboard PC. The onboard PC
consists of a robot model calculation, position estimator by Kalman
filter, and position controller. Afterward, the onboard PC transmits
commands to Arduino and Spinal with UART communication. Spinal also
receives self-attitude information from the IMU sensor and estimates the
attitude of the robot with a Kalman filter. Target PWM is output from
the attitude controller based on the estimated attitude values and
commands from the onboard PC. Moreover, Arduino receives commands from the onboard PC and sends angle commands to the servo motors of the coupling mechanism.
\subsection{Flight stability}
\begin{figure}[tb]
 \begin{center}
   \includegraphics[width=\columnwidth]{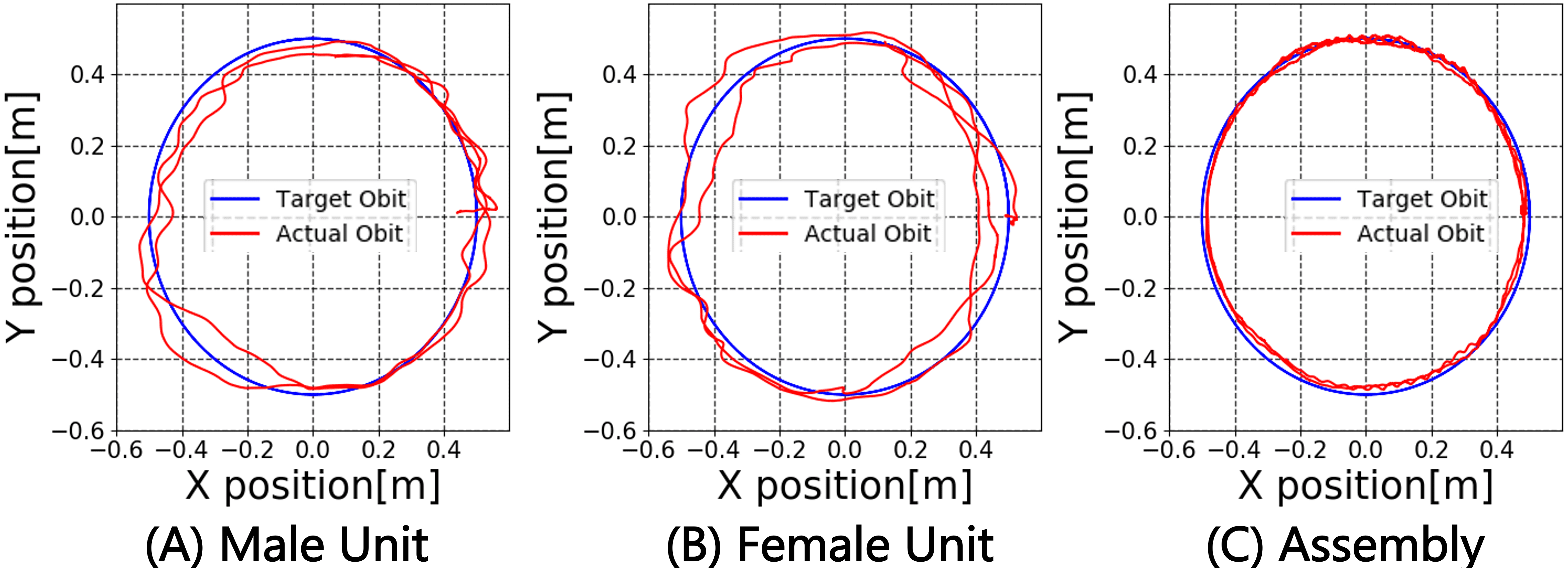}
 \end{center}
 \caption{Visualized tracking error. The blue trajectory represents the
 target trajectory, while the red trajectory represents the actual
 trajectory.
 }
   \label{figure:circle_track}
\end{figure}

We first conduct a circular trajectory tracking flight experiment in
order to verify the flight stability of both the unitary state and
the assembly state. In this experiment, a circular trajectory with a radius of
\SI{0.5}{m} at an altitude of \SI{1}{m} was adopted, and a robot circles
this obit twice in one minute. The result is shown
in \figref{figure:circle_track} and \tabref{table:circle_track}. Figure
1 visualizes the tracking error, while \tabref{table:circle_track} shows
the Root Mean Squared Error (RMSE) from the trajectory for each state.
From these results, it is found that both the under-actuated model
control in the unitary state and the fully-actuated model control in the
assembly state are possible. The control accuracy is inferior in the
unitary state compared to the assembly state, which is due
to not only the under-actuated model control but also the large moment of
inertia of the body despite its weight. However, the positional error
within this range can be
dealt with by the drogue on the female-side docking mechanism and the
proposed motion strategy, thus not posing significant issues.
\begin{table}[tb]
  \renewcommand{\arraystretch}{1.3}
  \caption{errors from the desired trajectory}
  \centering
 \begin{tabular}{|c|c|}
  \hline
  State& RSME\\
  \hline
  Male Unit& \SI{3.5}{cm}\\
  Female Female& \SI{4.1}{cm}\\
  Assembly &\SI{0.83}{cm}\\
  \hline
 \end{tabular}
 \label{table:circle_track}
\end{table}
\renewcommand{\arraystretch}{1.0}
\subsection{In-flight assembly and disassembly}
\subsubsection{Reliability of Aerial Assembly/Disassembly}
 To conduct an airborne coalescence experiment, the parameters in
 \eqref{equation:hash_1} and \eqref{equation:hash_2} must first be determined.
 The parameters $e^{\#1}_{y}$, $e^{\#1}_{z}$, $e^{\#1}_{\psi}$ in \eqref{equation:hash_1} are the criteria for determining the
 dangerous state, and therefore, smaller values of these parameters will
 result in more reliable merging, but will also increase the time
 required to complete the merging. Since the parameters $e^{\#2}_{x}$, $e^{\#2}_{y}$, $e^{\#2}_{z}$, $e^{\#2}_{\psi}$ in \eqref{equation:hash_2}
 are the criteria for docking certainty, there is also a tradeoff
 between certainty and time required to adjust these parameters. In this
 study, these values were adjusted through experiments on actual
 equipment, and the final values were determined as shown in
 \tabref{table:hash_cond}.\par
 After determining the parameters, we conducted 15 experiments for
 both aerial assembly and disassembly. The results indicated a success
 rate of \SI{86.7}{\%} (13 successful attempts) for assembly behavior and a
 \SI{100}{\%} success rate (15 successful attempts) for disassembly
 behavior. In addition to this, the time required for the merging
 behavior ranged from \SI{3}{s} to \SI{15}{s}, and there was a large
 variation. In two failed assembly experiments, the docking mechanism
 became stuck in a part of the airframe, making it impossible to
 transit from the ``Approach'' state to the ``Standby'' state,
 necessitating an emergency stop of the robot. As a solution to this
 issue, covering the entire unit body with a spherical guard would allow
 avoidance of contact between the docking mechanism and the airframe.
\begin{table}[tb]
  \renewcommand{\arraystretch}{1.3}
  \caption{parameter for motion strategy}
  \centering
 \begin{tabular}{|c|c|}
  \hline
  Parameter& Value\\
  \hline
  $e^{\#1}_{y}$ & $\pm$\SI{0.02}{m}\\
  $e^{\#1}_{z}$ & $\pm$\SI{0.02}{m}\\
  $e^{\#1}_{\psi}$ & $\pm$\SI{0.13}{rad}\\
  $e^{\#2}_{x}$ & $\pm$\SI{0.005}{m}\\
  $e^{\#2}_{y}$ & $\pm$\SI{0.01}{m}\\
  $e^{\#2}_{z}$ & $\pm$\SI{0.01}{m}\\
  $e^{\#2}_{\psi}$ & $\pm$\SI{0.01}{rad}\\
  $d^{st}$&\SI{0.6}{m} \\
  \hline
 \end{tabular}
 \label{table:hash_cond}
\end{table}
\renewcommand{\arraystretch}{1.0}
\subsubsection{Stability of System Switching}
Next, to evaluate the flight stability during the system switchover, we
conducted in-flight assembly and disassembly experiments with and without
the thrust transition method proposed in
Section. \secref{subsec:transition}. \par
\begin{figure}[tb]
 \begin{center}
   \includegraphics[width=0.8\columnwidth]{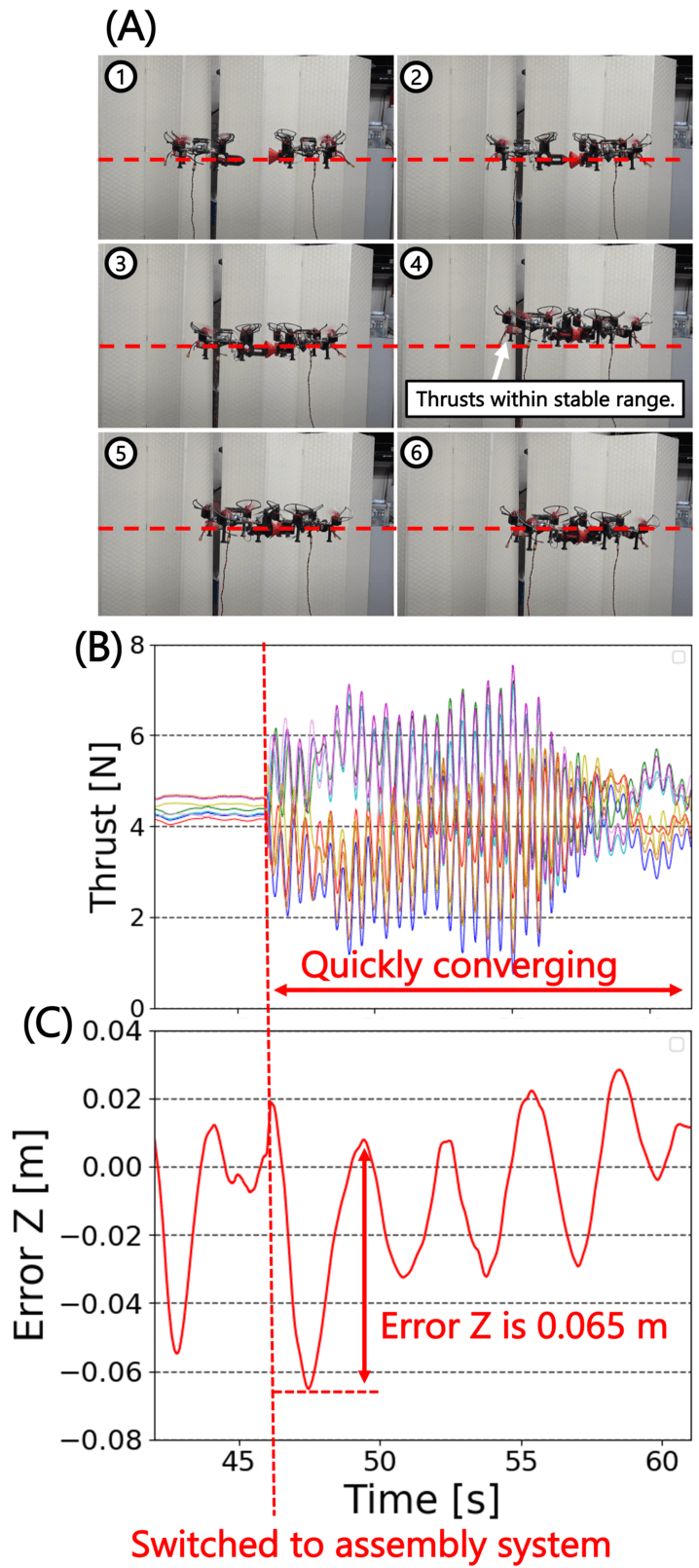}
 \end{center}
  \caption{Result of aerial assembly with presented transition
 process. (A)Spapshots of aerial assembly motion and the red line in the
 figure represents target altitude. (B)Plot of the thrusts of 8 rotors
 during the motion. (C)Plot of the Error of altitude $z$ during the motion.}
   \label{figure:assembly_trans}
\end{figure}
\begin{figure}[tb]
 \begin{center}
   \includegraphics[width=0.7\columnwidth]{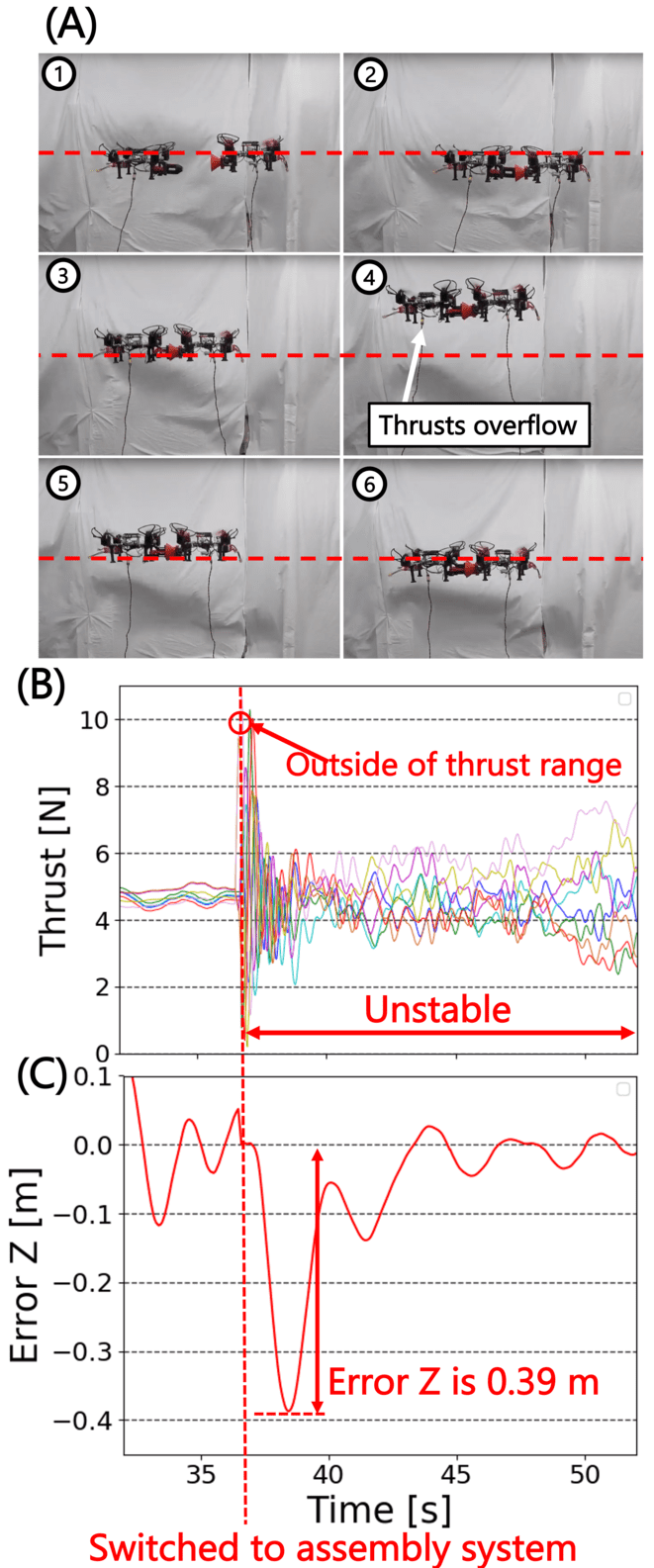}
 \end{center}
 \caption{Result of aerial assembly without presented transition
 process.(A)-(B) are the same as those in \figref{figure:assembly_trans}.}
   \label{figure:assembly_notrans}
\end{figure}

First, for the aerial assembly, the
results with the transition process are shown in \figref{figure:assembly_trans}, and the
results without the proposed thurst transition process are shown in
\figref{figure:assembly_notrans}. In the event of an increase in rotor
thrust during the system switch, the forces in the x and y directions
are offset; however, significant effects are apparent in the control of
the z direction. Therefore, both \figref{figure:assembly_trans} and \figref{figure:assembly_notrans} include plots of
rotor thrust displacement and z-directional position error.
Concerning the case of transition processing, it can be
observed from \figref{figure:assembly_trans} that stable aerial assembly and thrust
transitions are achieved. As shown in \figref{figure:assembly_trans}(B) and (C), the total
rotor thrust remains constant before and after the system switch, and
the subsequent ascent of robot is limited to approximately
\SI{0.065}{m}. Additionally, the control is rapidly stabilized following
the switchover. Note that, despite the transition process, there are still discrete changes in the
thrust of each rotor. However, these changes are due to alterations in
the allocation of thrust to each rotor and not to model errors, and they
have a negligible adverse effect on control performance.\par

On the contrary, regarding the experiment without
transition processing, \figref{figure:assembly_notrans}(B) shows that
the target rotor thrust changed abruptly due to model errors that occur
during the system switchover. The total rotor thrust increases by about \SI{5}{N} before and after the switchover, and for some rotors the
target thrust greatly exceeds the upper thrust limit. As a result, the
robot rose about \SI{0.39}{m} at a stretch immediately after assembly as
show in \figref{figure:assembly_notrans}(C), and
unstable control continued after that.  \par
These results indicate the efficacy of our proposed method for switching systems in
aerial assembly motion.\par
\begin{figure}[tb]
 \begin{center}
   \includegraphics[width=0.7\columnwidth]{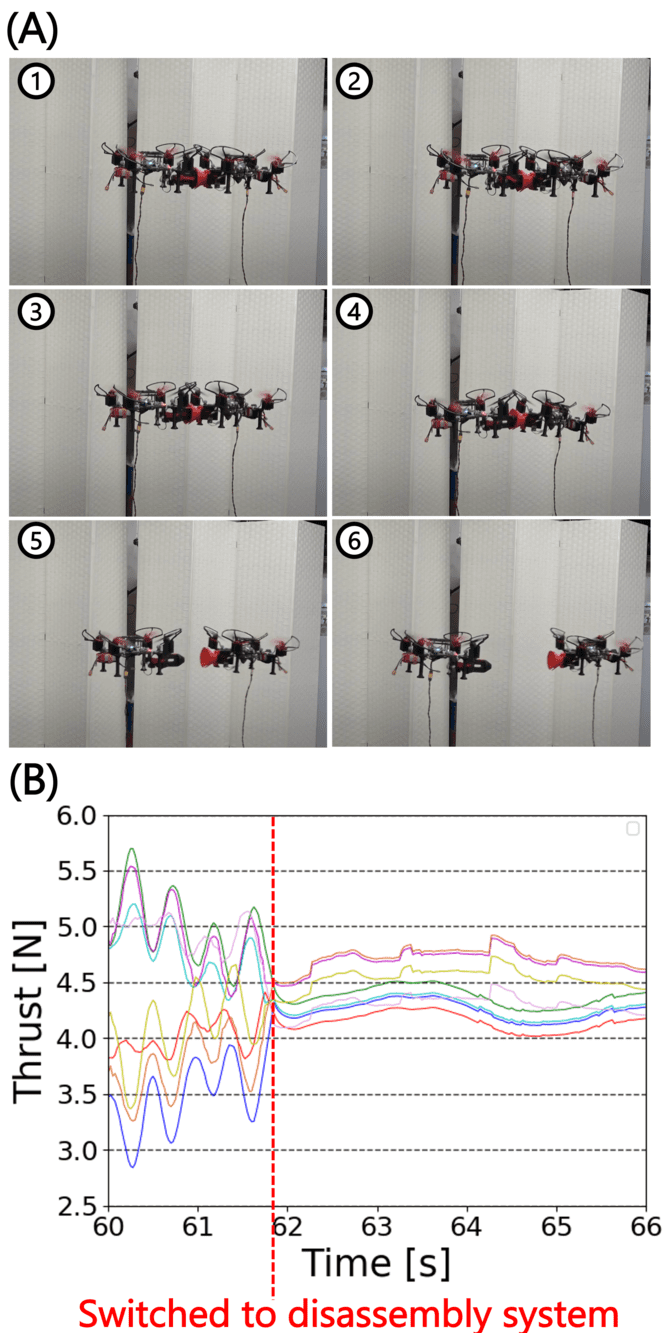}
 \end{center}
 \caption{Result of aerial disassembly with the presented transition
 process.(A)Spapshots of aerial disassembly motion.(B)Plot of the thrusts of 8 rotors
 during the motion.}
   \label{figure:disassembly_trans}
\end{figure}
\begin{figure}[tb]
 \begin{center}
   \includegraphics[width=0.7\columnwidth]{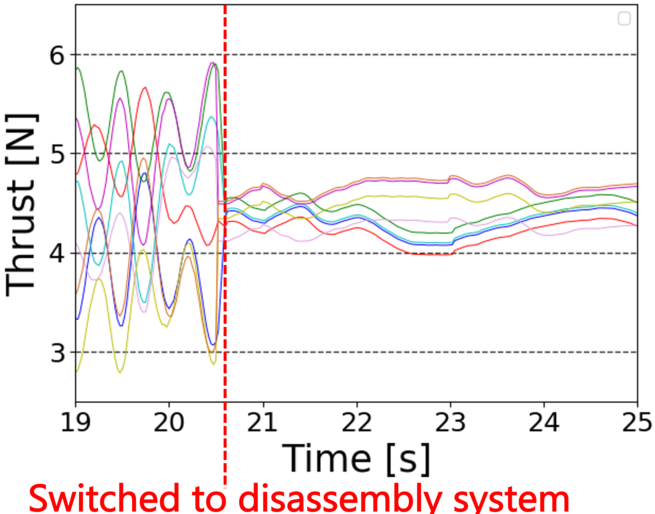}
 \end{center}
 \caption{Plot of the thrusts of 8 rotors during the disassembly without
 transition process.}
   \label{figure:disassembly_notrans}
\end{figure}

Next, regarding aerial disassembly, \figref{figure:disassembly_trans}
illustrates the experimental result of aerial disassembly with proposed
transition process. Moreover, \figref{figure:disassembly_notrans} illustrates the variation in
rotor thrust before and after mid-air disassembly when the transition
process was not applied. From these results, it was found that in aerial disassembly, it is
possible to safely switch the system with or without transition
processing. This is likely due to the fact that in the assembly state,
decentralized control is performed by two controllers, resulting in
model errors equivalent to those of two aircraft. In contrast, in the
unitary state, only model errors equivalent to those of one aircraft are
generated. As a result, in the unitary state, the effect of system switching on thrust is
presumed to be small.
\subsection{Peg Insertion}
\begin{figure}[tb]
 \begin{center}
   \includegraphics[width=\columnwidth]{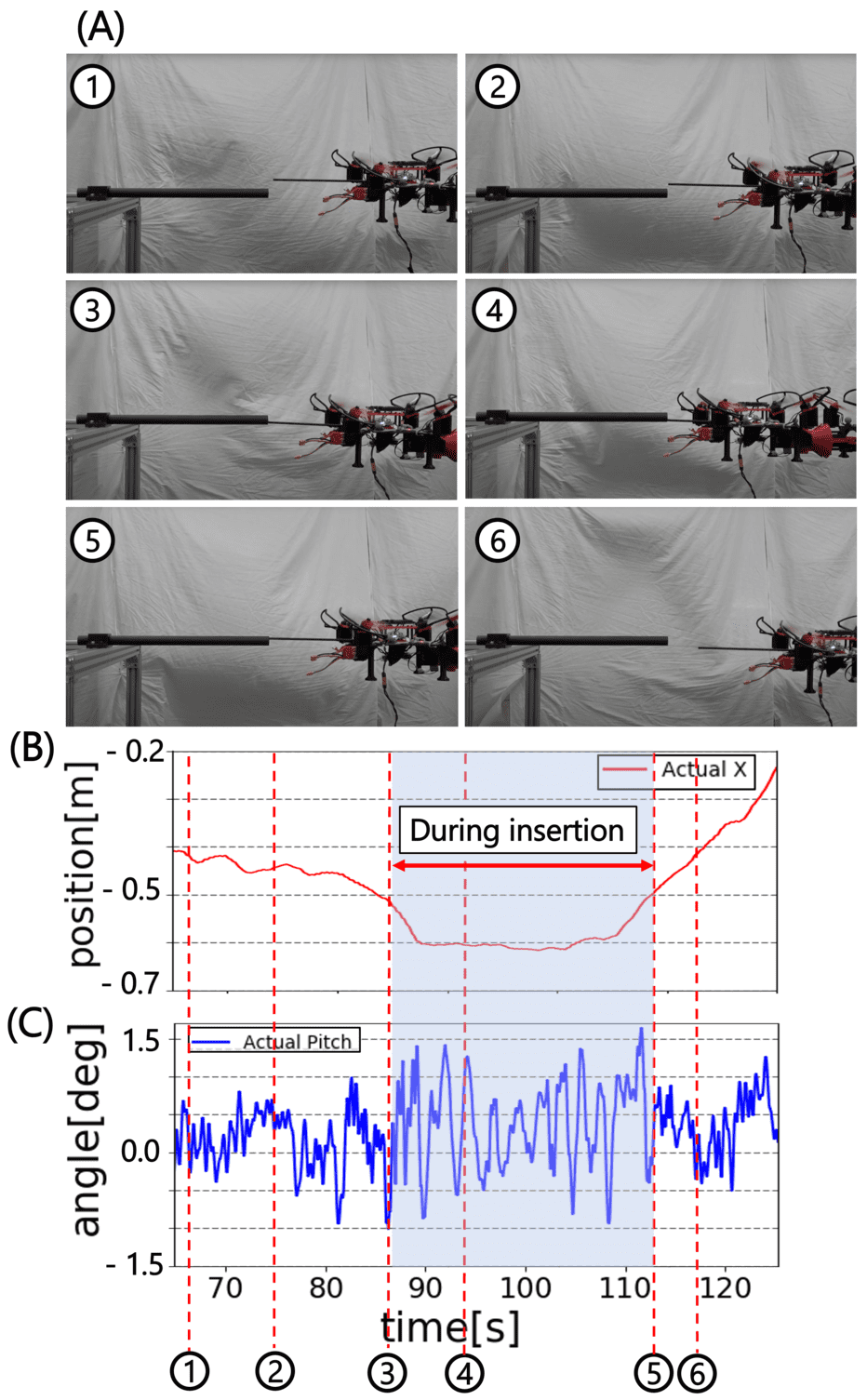}
 \end{center}
 \caption{Result of peg insertion experiment. (A)Snapshots of task
 execution. (B)Plot of the robot position $x$ during the task. (C)Plot
 of the robot's pitch angle $\phi$. The sky blue region in the figure is
 the duration of peg insertion.}
   \label{figure:peg}
\end{figure}

Next, we conducted an experiment to verify the aerial object
manipulation ability in the assembly state. First, we conducted an
experiment to insert a diameter of \SI{6}{mm} peg into a diameter of \SI{25}{mm} pipe as an analogy to drilling and Pic-and-Place
tasks.
 Due to the requirement of high position accuracy and independent
 control of translation and rotation for this task, the fully-actuated model
 control is necessary. Note that in this study, we focus on the control
 performance of the robot, and therefore, the task is performed by manual
 operation, not by automatic control based on motion planning. The
 experimental results are shown in \figref{figure:peg}.\par
 From these results, it was demonstrated that the robot can be
 maneuvered with an error of approximately $\pm$\SI{1}{cm}. Furthermore, as shown in \figref{figure:peg}(B),(C), the pitch angle of
 the robot was kept within approximately $\pm 1.5$ degrees
 during movement and hovering, demonstrating translational control can
 be achieved while suppressing the impact on the attitude.
\subsection{Valve Opening}
Finally, to demonstrate the expansion of achievable torque through
assembly of units, we conducted a valve opening experiment. An
industrial gate valve was used for the experiment, and end effectors
for valve operation was attached to the robot. These end effectors are
capable of passive expansion and contraction, and do not hinder
takeoff and landing. The results of the experiment are presented in
\figref{figure:peg}. From these results, it was demonstrated that
proposed robot TRADY
is capable of stable valve opening operations in the air. \figref{figure:peg}(B)
shows that a maximum torque of \SI{2.4}{Nm} is required during task execution
(actually, about \SI{2.2}{Nm} of torque is required to open this valve). Here,
it can be seen that by assembling units, the torque performance is
improved by approximately nine times with respect to the yaw direction
torque that can be exerted in unit state, which is a maximum of
\SI{0.28}{Nm}. This is due to the fact that the size of the body has increased by
assembling units and that torque realization using the horizontal
component of rotor thrusts is possible due to the fully-actuated model
control.
\begin{figure}[tb]
 \begin{center}
   \includegraphics[width=\columnwidth]{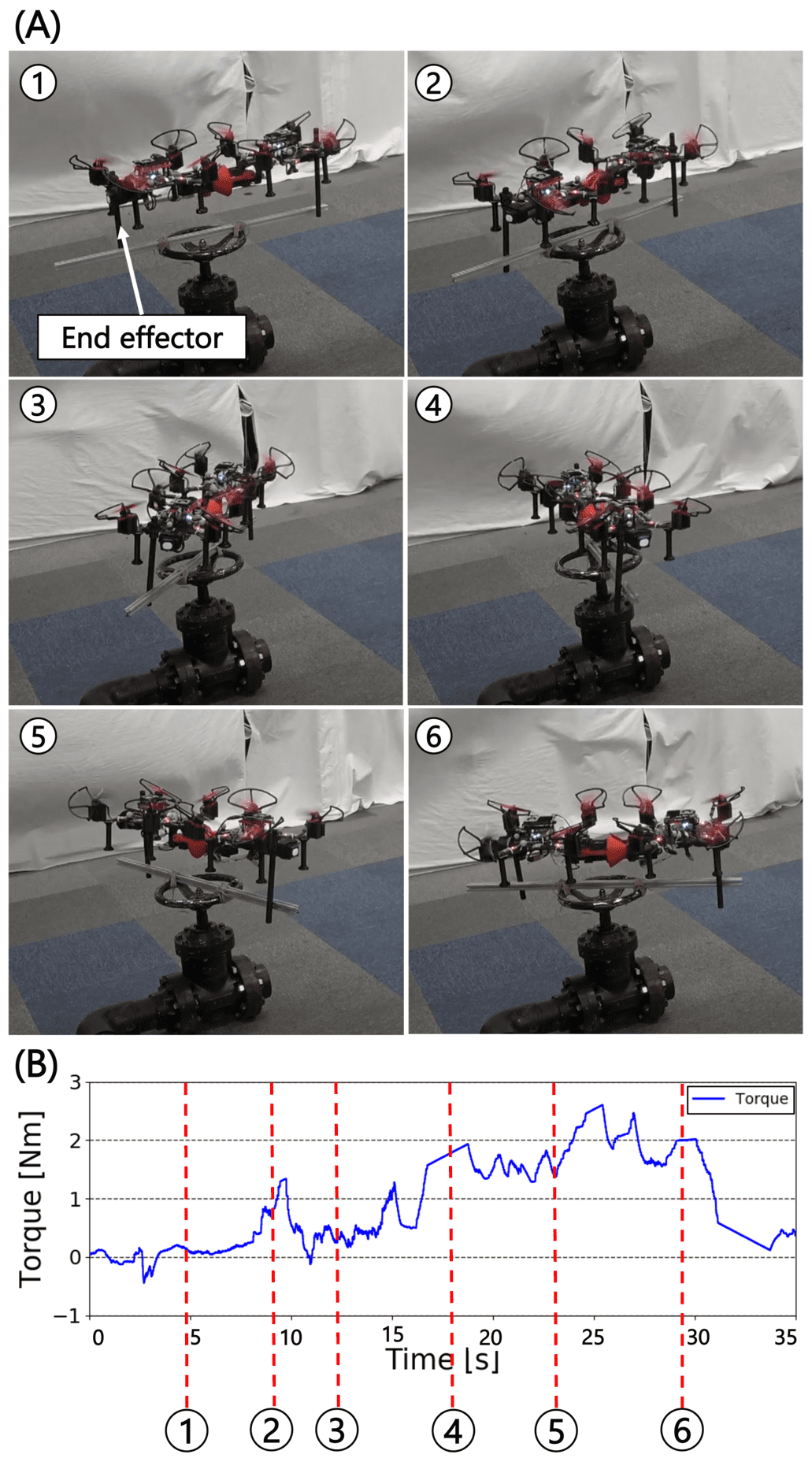}
 \end{center}
 \caption{Result of valve opening experiment. (A)Snapshots of task
 execution. (B)Plot of the robot's desired torque.}
   \label{figure:valve}
\end{figure}

\section{Conclusion}\label{sec:conclusion}
In this study, we have developed the configuration of the quadrotor unit
named TRADY that
possesses the ability to engage in aerial self-assembly and
self-disassembly. A noteworthy achievement of this work is that the
robot can perform both assembly and disassembly while seamlessly
transitioning between fully-actuated and under-actuated models. Building
upon the proposed design, control methodology, and motion strategy, we
conducted empirical experiments that demonstrated the robot's ability to
perform stable assembly/disassembly movements, as well as execute
various aerial manipulation tasks. During the assembly/disassembly
experiment, we established that the proposed robot can successfully
complete assembly/disassembly movements at a rate of 90\%, while also
observing that the proposed thrust transition method can suppress
instability during switching. Additionally, in the task
execution experiment, we determined that the robot in the assembly state
can exercise independent control over translation and rotation, while
also generating nine times the torque compared to the unitary state.\par
The pivotal concern that remains in this study is that TRADY, while in
the assembly state, is fully-actuated but not omni-directional, which
restricts its ability to hover at a significantly tilted posture. As a
future prospect, we intend to design a new docking mechanism equipped
with joints that will enable the robot to alter rotor directions after
assembly. This will expand the robot's controllability in a more
significant manner. Furthermore, expanding the system by utilizing three
or more units remains a future challenge. In cases involving three or
more units, multiple combinations are possible, allowing for the
selection of appropriate units based on the task at hand.
\FloatBarrier

% \appendices

% % you can choose not to have a title for an appendix
% % if you want by leaving the argument blank
% \section{}

% % use section* for acknowledgment
% \section*{Acknowledgment}


% \ifCLASSOPTIONcaptionsoff
%   \newpage
% \fi



\bibliographystyle{IEEEtran}
% \bibliography{IEEEabrv,bib}

\begin{thebibliography}{10}
\providecommand{\url}[1]{#1}
\csname url@samestyle\endcsname
\providecommand{\newblock}{\relax}
\providecommand{\bibinfo}[2]{#2}
\providecommand{\BIBentrySTDinterwordspacing}{\spaceskip=0pt\relax}
\providecommand{\BIBentryALTinterwordstretchfactor}{4}
\providecommand{\BIBentryALTinterwordspacing}{\spaceskip=\fontdimen2\font plus
\BIBentryALTinterwordstretchfactor\fontdimen3\font minus
  \fontdimen4\font\relax}
\providecommand{\BIBforeignlanguage}[2]{{%
\expandafter\ifx\csname l@#1\endcsname\relax
\typeout{** WARNING: IEEEtran.bst: No hyphenation pattern has been}%
\typeout{** loaded for the language `#1'. Using the pattern for}%
\typeout{** the default language instead.}%
\else
\language=\csname l@#1\endcsname
\fi
#2}}
\providecommand{\BIBdecl}{\relax}
\BIBdecl

\bibitem{Floreano2015}
D.~Floreano and R.~J. Wood, ``Science, technology and the future of small
  autonomous drones,'' \emph{Nature}, vol. 521, pp. 460--466, 2015.

\bibitem{kumar2012}
V.~Kumar and N.~Michael, ``Opportunities and challenges with autonomous micro
  aerial vehicles,'' \emph{The International Journal of Robotics Research},
  vol.~31, no.~11, pp. 1279--1291, 2012.

\bibitem{Bonatti2020}
R.~Bonatti, Y.~Zhang, S.~Choudhury, W.~Wang, and S.~Scherer, ``Autonomous drone
  cinematographer: Using artistic principles to create smooth, safe,
  occlusion-free trajectories for aerial filming,'' in \emph{Proceedings of the
  2018 international symposium on experimental robotics}.\hskip 1em plus 0.5em
  minus 0.4em\relax Springer, 2020, pp. 119--129.

\bibitem{Sewer}
F.~Chataigner, ``Arsi: an aerial robot for sewer inspection.'' \emph{Advances
  in Robotics Research: From Lab to Market. Springer, Cham, 2020}, pp.
  249--274, 2020.

\bibitem{Michael2012}
N.~Michael, S.~Shen, K.~Mohta, Y.~Mulgaonkar, V.~Kumar, K.~Nagatani, Y.~Okada,
  S.~Kiribayashi, K.~Otake, K.~Yoshida, K.~Ohno, E.~Takeuchi, and S.~Tadokoro,
  ``Collaborative mapping of an earthquake-damaged building via ground and
  aerial robots,'' \emph{Journal of Field Robotics}, vol.~29, no.~5, pp.
  832--841, 2012.

\bibitem{Doitsidis2012}
L.~Doitsidis, S.~Weiss, A.~Renzaglia, M.~W. Achtelik, E.~Kosmatopoulos,
  R.~Siegwart, and D.~Scaramuzza, ``Optimal surveillance coverage for teams of
  micro aerial vehicles in gps-denied environments using onboard vision,''
  \emph{Autonomous Robots}, vol.~33, pp. 173--188, 2012.

\bibitem{figure_8}
Y.~Bai and S.~Gururajan, ``Evaluation of a baseline controller for autonomous
  “figure-8” flights of a morphing geometry quadcopter: Flight performance,''
  \emph{Drones}, 2019.

\bibitem{morphing_quad}
D.~Falanga, K.~Kleber, S.~Mintchev, D.~Floreano, and D.~Scaramuzza, ``The
  foldable drone: A morphing quadrotor that can squeeze and fly,'' \emph{IEEE
  Robotics and Automation Letters}, vol.~4, no.~2, pp. 209--216, 2019.

\bibitem{origami}
D.~Yang, S.~Mishra, D.~M. Aukes, and W.~Zhang, ``Design, planning, and control
  of an origami-inspired foldable quadrotor,'' in \emph{2019 American Control
  Conference (ACC)}, 2019, pp. 2551--2556.

\bibitem{n_zhao}
N.~Zhao, Y.~Luo, H.~Deng, and Y.~Shen, ``The deformable quad-rotor: Design,
  kinematics and dynamics characterization, and flight performance
  validation,'' in \emph{2017 IEEE/RSJ International Conference on Intelligent
  Robots and Systems (IROS)}, 2017, pp. 2391--2396.

\bibitem{Bucki}
N.~Bucki, J.~Tang, and M.~W. Mueller, ``Design and control of a
  midair-reconfigurable quadcopter using unactuated hinges,'' \emph{IEEE
  Transactions on Robotics}, vol.~39, no.~1, pp. 539--557, 2023.

\bibitem{Voliro}
``The voliro omniorientational hexacopter: An agile and maneuverable
  tiltable-rotor aerial vehicle,'' \emph{IEEE Robotics and Automation
  Magazine}, vol.~25, pp. 34--44, 12 2018.

\bibitem{khamseh2018}
H.~B. Khamseh, F.~Janabi-Sharifi, and A.~Abdessameud, ``Aerial manipulation―a
  literature survey,'' \emph{Robotics and Autonomous Systems}, vol. 107, pp.
  221--235, 2018.

\bibitem{Mellinger2011}
D.~Mellinger, Q.~Lindsey, M.~Shomin, and V.~Kumar, ``Design, modeling,
  estimation and control for aerial grasping and manipulation,'' in \emph{2011
  IEEE/RSJ International Conference on Intelligent Robots and Systems}, 2011,
  pp. 2668--2673.

\bibitem{Heredia2014}
G.~Heredia, A.~Jimenez-Cano, I.~Sanchez, D.~Llorente, V.~Vega, J.~Braga,
  J.~Acosta, and A.~Ollero, ``Control of a multirotor outdoor aerial
  manipulator,'' in \emph{2014 IEEE/RSJ International Conference on Intelligent
  Robots and Systems}, 2014, pp. 3417--3422.

\bibitem{hydrus}
M.~Zhao, K.~Kawasaki, K.~Okada, and M.~Inaba, ``Transformable multirotor with
  two-dimensional multilinks: modeling, control, and motion planning for aerial
  transformation,'' \emph{Advanced Robotics}, vol.~30, no.~13, pp. 825--845,
  2016.

\bibitem{dragon}
M.~Zhao, F.~Shi, T.~Anzai, K.~Okada, and M.~Inaba, ``Online motion planning for
  deforming maneuvering and manipulation by multilinked aerial robot based on
  differential kinematics,'' \emph{IEEE Robotics and Automation Letters},
  vol.~5, no.~2, pp. 1602--1609, 2020.

\bibitem{Lasdra}
H.~Yang, S.~Park, J.~Lee, J.~Ahn, D.~Son, and D.~Lee, ``Lasdra: Large-size
  aerial skeleton system with distributed rotor actuation,'' in \emph{2018 IEEE
  International Conference on Robotics and Automation (ICRA)}, 2018, pp.
  7017--7023.

\bibitem{Zhao2022}
M.~Zhao, K.~Okada, and M.~Inaba, ``Versatile articulated aerial robot dragon:
  Aerial manipulation and grasping by vectorable thrust control,''
  \emph{International Journal of Robotics Research}, 2022.

\bibitem{Naldi2015}
R.~Naldi, F.~Forte, A.~Serrani, and L.~Marconi, ``Modeling and control of a
  class of modular aerial robots combining under actuated and fully actuated
  behavior,'' \emph{IEEE Transactions on Control Systems Technology}, vol.~23,
  no.~5, pp. 1869--1885, 2015.

\bibitem{Granger}
K.~Garanger, J.~Epps, and E.~Feron, ``Modeling and experimental validation of a
  fractal tetrahedron uas assembly,'' in \emph{2020 IEEE Aerospace Conference},
  2020, pp. 1--11.

\bibitem{Xu2021}
J.~Xu, D.~S. D'Antonio, and D.~Saldana, ``H-modquad: Modular multi-rotors with
  4, 5, and 6 controllable dof,'' vol. 2021-May.\hskip 1em plus 0.5em minus
  0.4em\relax Institute of Electrical and Electronics Engineers Inc., 2021, pp.
  190--196.

\bibitem{Flightarray1}
R.~Oung, A.~Ramezani, and R.~D'Andrea, ``Feasibility of a distributed flight
  array,'' in \emph{Proceedings of the 48h IEEE Conference on Decision and
  Control (CDC) held jointly with 2009 28th Chinese Control Conference}, 2009,
  pp. 3038--3044.

\bibitem{Flightarray2}
R.~Oung, F.~Bourgault, M.~Donovan, and R.~D'Andrea, ``The distributed flight
  array,'' in \emph{2010 IEEE International Conference on Robotics and
  Automation}, 2010, pp. 601--607.

\bibitem{Saldana2018}
D.~Saldana, B.~Gabrich, G.~Li, M.~Yim, and V.~Kumar, ``Modquad: The flying
  modular structure that self-assembles in midair.''\hskip 1em plus 0.5em minus
  0.4em\relax Institute of Electrical and Electronics Engineers Inc., 9 2018,
  pp. 691--698.

\bibitem{moddessemble}
D.~Saldana, P.~M. Gupta, and V.~Kumar, ``Design and control of aerial modules
  for inflight self-disassembly,'' \emph{IEEE Robotics and Automation Letters},
  vol.~4, pp. 3402--3409, 10 2019.

\bibitem{Hara2014}
I.~O'Hara, J.~Paulos, J.~Davey, N.~Eckenstein, N.~Doshi, T.~Tosun, J.~Greco,
  J.~Seo, M.~Turpin, V.~Kumar, and M.~Yim, ``Self-assembly of a swarm of
  autonomous boats into floating structures,'' in \emph{2014 IEEE International
  Conference on Robotics and Automation (ICRA)}, 2014, pp. 1234--1240.

\bibitem{Yanagimura}
K.~Yanagimura, K.~Ohno, Y.~Okada, E.~Takeuchi, and S.~Tadokoro, ``Hovering of
  mav by using magnetic adhesion and winch mechanisms,'' in \emph{2014 IEEE
  International Conference on Robotics and Automation (ICRA)}, 2014, pp.
  6250--6257.

\bibitem{Ryll2016}
M.~Ryll, D.~Bicego, and A.~Franchi, ``Modeling and control of fast-hex: A
  fully-actuated by synchronized-tilting hexarotor,'' in \emph{2016 IEEE/RSJ
  International Conference on Intelligent Robots and Systems (IROS)}.\hskip 1em
  plus 0.5em minus 0.4em\relax IEEE, 2016, pp. 1689--1694.

\bibitem{Park2016}
S.~Park, J.~Her, J.~Kim, and D.~Lee, ``Design, modeling and control of
  omni-directional aerial robot,'' vol. 2016-November.\hskip 1em plus 0.5em
  minus 0.4em\relax Institute of Electrical and Electronics Engineers Inc., 11
  2016, pp. 1570--1575.

\bibitem{Tandale}
M.~D. Tandale, R.~Bowers, and J.~Valasek, ``Trajectory tracking controller for
  vision-based probe and drogue autonomous aerial refueling,'' \emph{Journal of
  Guidance, Control, and Dynamics}, vol.~29, no.~4, pp. 846--857, 2006.

\bibitem{FRAVOLINI2004611}
\BIBentryALTinterwordspacing
M.~L. Fravolini, A.~Ficola, G.~Campa, M.~R. Napolitano, and B.~Seanor,
  ``Modeling and control issues for autonomous aerial refueling for uavs using
  a probe^^e2^^80^^93drogue refueling system,'' \emph{Aerospace Science and
  Technology}, vol.~8, no.~7, pp. 611--618, 2004. [Online]. Available:
  \url{https://www.sciencedirect.com/science/article/pii/S1270963804000744}
\BIBentrySTDinterwordspacing

\bibitem{singularity}
M.~Zhao, T.~Anzai, K.~Okada, K.~Kawasaki, and M.~Inaba, ``Singularity-free
  aerial deformation by two-dimensional multilinked aerial robot with 1-dof
p  vectorable propeller,'' \emph{IEEE Robotics and Automation Letters}, vol.~6,
  pp. 1367--1374, 4 2021.

\bibitem{Convex}
P.~Bosscher, A.~Riechel, and I.~Ebert-Uphoff, ``Wrench-feasible workspace
  generation for cable-driven robots,'' \emph{IEEE Transactions on Robotics},
  vol.~22, no.~5, pp. 890--902, 2006.

\bibitem{Isres}
N.~Documentation, ``Nlopt algorithms,''
  \url{https://nlopt.readthedocs.io/en/latest/NLopt_Algorithms/#isres-improved-stochastic-ranking-evolution-strategy}.

\bibitem{Lee2010}
T.~Lee, M.~Leok, and N.~H. McClamroch, ``Geometric tracking control of a
  quadrotor uav on se(3),'' in \emph{49th IEEE Conference on Decision and
  Control (CDC)}, 2010, pp. 5420--5425.

\bibitem{LQI}
P.~Young and J.~Willems, ``An approach to the linear multivariable
  servomechanism problem,'' \emph{International Journal of Control}, vol.~15,
  06 1972.

\end{thebibliography}
% \printbibliography


% \end{thebibliography}


% \end{IEEEbiography}



\end{document}